\documentclass[bst/sn-mathphys,Numbered]{sn-jnl}

\usepackage{graphicx}%
\usepackage{multirow}%
\usepackage{amsmath,amssymb,amsfonts}%
\usepackage{amsthm}%
\usepackage{mathrsfs}%
\usepackage[title]{appendix}%
\usepackage{xcolor}%
\usepackage{textcomp}%
\usepackage{manyfoot}%
\usepackage{booktabs}%
\usepackage{algorithm}%
\usepackage{algorithmicx}%
\usepackage{algpseudocode}%
\usepackage{listings}%
\usepackage[utf8]{inputenc}
\usepackage{glossaries}
\usepackage{hyperref}
\usepackage{cleveref}
\usepackage{amsthm}
\usepackage{mwe}
\usepackage{wrapfig}
\Crefname{figure}{Fig.}{Figs.}
\Crefname{table}{Tab.}{Tabs.}
\Crefname{section}{Sec.}{Secs.}
\Crefname{definition}{Def.}{Defs.}
\Crefname{equation}{Eqn.}{Eqns.}
\usepackage{tikz}
\usepackage{tkz-tab}
\usetikzlibrary{positioning}
\usepackage{caption}
\usepackage{subcaption}
\usepackage{latexsym}
\usepackage{amssymb}
\usepackage{amsmath}
\usepackage{subcaption} 
\usepackage{enumitem}
\usepackage[numbers]{natbib}
\usetikzlibrary{arrows,calc,decorations.markings,math,arrows.meta}

\usepackage[normalem]{ulem} 
\usepackage{dirtytalk} 
\DeclareMathOperator{\curl}{curl}
\DeclareMathOperator{\Div}{div}

\DeclareMathOperator*{\argmin}{arg\,min}
\newacronym{nn}{NN}{neural network}
\newacronym{lstm}{LSTM}{long short-term memory network}
\newacronym{rom}{ROM}{reduced-order model}
\newacronym{cnn}{CNN}{convolutional neural network}
\newacronym{hym}{HYM}{hybrid modeling}
\newacronym{pinn}{PINN}{physics-informed neural network}
\newacronym{rnn}{RNN}{recurrent neural network}
\newacronym{gru}{GRU}{gated recurrent unit}
\newacronym{cru}{CRU}{continuous recurrent unit}
\newacronym{ode}{ODE}{ordinary differential equation}
\newacronym{pde}{PDE}{partial differential equation}
\newacronym{sde}{SDE}{stochastic differential equation}

\theoremstyle{thmstyleone}%

\theoremstyle{thmstyletwo}%

\theoremstyle{thmstylethree}%
\newtheorem{definition}{Definition}%

\renewcommand{\glossarysection}[2][]{}
\makeglossaries
\begin{document}

\title[Hybrid Modeling Design Patterns]{Hybrid Modeling Design Patterns}


\author*[1]{\fnm{Maja} \sur{Rudolph}}\email{maja.rudolph@us.bosch.com}

\author*[2,3]{\fnm{Stefan} \sur{Kurz}}\email{stefan.kurz2@de.bosch.com}

\author*[2]{\fnm{Barbara} \sur{Rakitsch}}\email{barbara.rakitsch@de.bosch.com}

\affil[1]{\orgdiv{Bosch Center for AI}, 
\orgaddress{\city{Pittsburgh}, \state{PA}, \country{USA}}}

\affil[2]{\orgdiv{Bosch Center for AI}, 
\orgaddress{\city{Renningen}, \country{Germany}}}
\affil[3]{\orgdiv{University of Jyväskylä},
\orgaddress{\country{Finland}}}

\abstract{Design patterns provide a systematic way to convey solutions to recurring modeling challenges. This paper introduces design patterns for hybrid modeling, an approach that combines modeling based on first principles with data-driven modeling techniques. While both approaches have complementary advantages there are often multiple ways to combine them into a hybrid model, and the appropriate solution will depend on the problem at hand. In this paper, we provide four base patterns that can serve as blueprints for combining data-driven components with domain knowledge into a hybrid approach. In addition, we also present two composition patterns that govern the combination of the base patterns into more complex hybrid models. Each design pattern is illustrated by typical use cases from application areas such as climate modeling, engineering, and physics.}


\maketitle

\section{Introduction}
Models play a crucial role in the scientific process by providing a representation of complex systems, processes, and phenomena. Models help scientists to make predictions, test hypotheses, and gain a deeper understanding of the behavior of these systems. By using mathematical models, such as physical, statistical, or simulation models, scientists can study the relationships between variables, estimate uncertainties, and explore scenarios without having to perform expensive or dangerous experiments. In this way, models serve as a powerful tool for advancing our knowledge and understanding of the world, and for solving real-world problems in fields such as medicine, engineering, and environmental science.\par
Traditionally, models are derived from first principles and encode domain knowledge such as physical laws or physical constraints. Such models emerge from the scientific process through a combination of observation, experimentation, and theoretical analysis. After careful observation of natural phenomena, scientists form hypotheses and theories to explain the observed behavior. These theories are then tested through experiments and compared with existing knowledge and models. If a theory withstands experimental scrutiny and provides accurate predictions, it may become accepted as a law or constraint. Models based on first principles are data-efficient, causal, lead to explainable predictions, are often more reliable than data-driven models since the underlying theory has been validated, and predictions will generalize to other deployment regimes as long as the underlying assumptions of the model still hold.\par
Data-driven models, on the other hand, are a type of modeling approach that relies on large data sets to identify patterns and correlations in the data that can be used to make predictions or classifications. These models are often used in fields where the underlying physical processes are too complex to model by first-principles. Data-driven models are typically developed using machine learning techniques such as neural networks. These models can be trained on large data sets of labeled and in some cases unlabeled data and can then be used to make predictions or classifications on new data. Data-driven models have shown promise in a wide range of applications, including image and speech recognition, natural language processing, and predictive modeling in finance and healthcare.\par
Hybrid models combine the strengths of both data-driven and first-principle based models, and can be useful in situations where neither approach alone is sufficient \citep{karpatne2017theory,von2019informed,willard2022integrating,kurz2022hybrid}. 
For example, mechanistic models are based on first principles and describe a hypothesized causal process between variables \cite{hilborn2013ecological}.
While they can provide a deep understanding of the underlying physics or biology of a system, they may not always capture all of the relevant details or interactions, leading to inaccuracies. 
On the other hand, data-driven models can accurately capture complex relationships in large data sets, but may not be able to explain the underlying mechanisms or provide insight into how the system behaves under new conditions. Hybrid models can combine the strengths of both approaches, allowing for more accurate and interpretable predictions even in complex systems with incomplete understanding of the underlying mechanisms.\par
Hybrid modeling is challenging because it requires expertise in both first-principle-based modeling and data-driven modeling, as well as knowledge of how to integrate the two approaches effectively. It can be difficult to determine the appropriate level of complexity for each component of the hybrid model and to ensure that the different components are compatible with each other. In particular, hybrid modeling requires careful consideration of the trade-offs between accuracy, complexity, interpretability, and scalability, which can be difficult to optimize.\par
Validating and verifying a hybrid model presents another challenge. Its data-driven and physics-based components may contribute different sources of uncertainty and error which need to be handled with care. For these reasons, designing and implementing a hybrid model requires careful consideration of the strengths and weaknesses of each modeling approach and a thorough understanding of the system being modeled.\par
The applications of hybrid modeling are incredibly diverse, spanning a wide range of fields and industries. From molecular modeling in drug discovery \citep{stokes2020deep}, to simulation tasks in climate \citep{beucler2019achieving} and earth science \citep{reichstein2019deep} and engineering, to modeling sensor data for virtual sensing \citep{liu2009virtual}, hybrid modeling is used in many domains to address unique and complex challenges.\par
This diversity of applications means that there is a need for solutions that can be applied more broadly, rather than being specific to one particular domain. Developing such approaches requires a focus on abstraction and generalization, so that solutions can be formulated at a higher level of abstraction that can be applied across multiple domains. This is where the concept of {\em design patterns} comes in, as a way to formalize recurring modeling challenges and to distill useful solution approaches that generalize across applications.\par
Formalizing solutions to recurring modeling challenges into hybrid modeling design patterns provides several benefits. First, it allows for the sharing of knowledge and expertise across application domains, which can lead to faster progress and innovation. Second, it facilitates the development of standardized tools and techniques for hybrid modeling, which can improve the efficiency and reliability of the modeling process. Third, it can help identify common challenges and limitations in hybrid modeling, which can guide future research directions and advance the field as a whole. Overall, the use of hybrid modeling design patterns can improve the accessibility, efficiency, and effectiveness of hybrid modeling across a wide range of applications.\par
\section{Background}
In this background section, we introduce modeling and then review both the first-principles-based as well as the data-driven perspective on modeling.
\subsection{Computational models} 
\label{sec:model}
The goal of hybrid modeling is to build a computational model for a system of interest. A computational model is a set of computations that are applied to an input to produce an output. The model of a system can be used to make predictions about how the system would react to certain inputs or to study how the system behaves under certain conditions. Alternatively, the model can be used to simulate the system. Models typically {\em approximate} the behavior of the underlying system, which might be too complex to model more accurately.\par
An computational model is of the form 
\begin{equation}\label{eq:comp_model}
    y = u(x).
\end{equation}
The inputs $x$ are manipulated by a function $u$ to produce the outputs $y$. The functional form of $u$ will depend on the model type. 
We distinguish between two different model types: The first type is models based on first principles, for example from physics. These are sometimes also called scientific models, and we often call them physical models. The second type of model is data-driven. Here one uses data to find a model within a class of functions that best explains the data. This function is then used as a model, e.g.~to make predictions.
\subsection{Modeling from first principles}
When modeling from first principles, the choice of $u$ is derived using scientific reasoning. There is a justification for both the functional form of $u$ and for the choice of its parameters. For this reason, these models are often called models based on first-principles, mechanistic models, physics-based models or science-based models.\par
For example, laws of physics, such as Newton's laws of motion and the law of conservation of energy, emerged from centuries of observation and experimentation in the field of mechanics. These laws provide a mathematical framework for understanding and predicting the behavior of physical systems, and have been tested and confirmed through numerous experiments. Similarly, in chemistry, conservation laws, such as the law of conservation of mass, emerged from the study of chemical reactions and provide a fundamental understanding of the behavior of chemical systems.\par
From a mathematical point of view, scientific models frequently take the form of {\em algebraic models}, {\em ordinary differential equations} (ODEs) or {\em partial differential equations} (PDEs), respectively.
\paragraph*{Algebraic models}
An algebraic mathematical model is a type of mathematical model that uses algebraic equations or functions to represent a real-world situation or system. In an algebraic model, the relationships between the variables are often represented using equations that involve elementary mathematical operations and functions.\par
One example is the equation for the trajectory of a stone that is vertically thrown in the air, where air resistance is neglected. The height $u(t)$ over ground as a function of time $t\ge 0$ is
\begin{equation}\label{eq:stone_example}
u(t) = -0.5gt^2 + v_0t + h_0\,,
\end{equation}
where $h_0$ is the initial height, $v_0$ the initial velocity and $g$ the gravitational constant.\par
From a computational perspective, this model could be utilized to compute -- for a given instance $t_1$ -- the height at this instance, $h_1=u(t_1)$.
\paragraph*{Ordinary differential equations (ODEs)}
A more involved model class are differential equations. An ODE is a type of differential equation that involves only one independent variable, usually time $t$, and its derivatives.\par
ODE models are particularly useful for systems that involve dynamic behavior, where the behavior of the system changes over time in response to internal or external factors. In an ODE model, the behavior of a system is represented using one or more ODEs that describe the rates of change of the system's variables. The ODEs can be used to predict how the system will evolve over time, based on its initial conditions and the values of its parameters.\par
Solving an ODE involves finding a mathematical expression that describes the behavior of the system as a function of time. This can be done using various analytical or numerical methods, depending on the complexity of the system and the accuracy of the desired solution. A closed form solution of an ODE yields an algebraic model. For example, the algebraic model \eqref{eq:stone_example} is a solution to the ODE
\[
\frac{\mathrm{d}^2u(t)}{\mathrm{d}u^2}=-g\,,
\]
subject to given initial conditions. This is just Newton's law, the first-principle based model that underlies the mechanistic model \eqref{eq:stone_example}.\par
Once a solution has been obtained, it can be used to predict the behavior of the system under different conditions or to design interventions to achieve a desired outcome.\par
In the following, we will consider three additional ODE models that will serve as recurring examples throughout the remainder of the paper.
\begin{enumerate}
\item
Let us start with the ODE of an {\em harmonic oscillator}
\begin{equation}
    \frac{\mathrm{d}^2u(t)}{\mathrm{d}t^2} = -u(t)\,,
\end{equation}
where $u(t)$ yields the normalized displacement at normalized time $t$. The normalization is with respect to some reference displacement $s_0$ and the oscillatory period $T$, respectively. For a spring-mass system with mass $m$ and spring constant $k$ the oscillatory period is $T=\sqrt{m/k}$. The model gets more interesting if a nonlinear damping term is added,
\begin{equation}\label{eq:van_der_pol}
    \frac{\mathrm{d}^2u(t)}{\mathrm{d}t^2} = -u(t)+\mu\frac{\mathrm{d}u(t)}{\mathrm{d}t}\bigl(1-u(t)^2\bigr)\,,
\end{equation}
where the positive real parameter $\mu$ determines the amount of nonlinear damping. Equation \eqref{eq:van_der_pol} is the {\em Van der Pol equation}, which exhibits a number of interesting nonlinear phenomena, such as relaxation oscillations~\citep{grasman1987asymptotic}.
\item
The {\em Lodtka-Volterra} equations are used to model the population dynamics of two interacting species of a predator and its prey. The population density of prey is $u(t)$ and the population density of predators is $w(t)$. The population dynamics is modeled by the nonlinear {\em system of ODEs}
\begin{equation}
\frac{\mathrm{d}u(t)}{\mathrm{d}t} = \alpha u(t) - \beta u(t)w(t)\,,\qquad
\frac{\mathrm{d}w(t)}{\mathrm{d}t} = \delta u(t)w(t) - \gamma w(t)\,,
\end{equation}
with positive real parameters $\alpha$, $\beta$, $\gamma$, and $\delta$ determining the self and mutual interactions of the two species.
\item
The simplest standard model for a dynamical system with several degrees of freedom is a {\em linear system of ODEs}, of the form
\begin{equation}\label{eq:linear_system_odes}
\frac{\mathrm{d}u(t)}{\mathrm{d}t}=f\bigl(u(t),t;\theta\bigr)\,,
\end{equation}
where $u(t)\in\mathbb{R}^n$ describes the state of the system at time $t$, a point in an $n$-dimensional state space. Herein, $\theta\in\mathbb{R}^p$ is a $p$-dimensional parameter vector that admits calibrating the model. Given an initial condition $u(t_0)$ at time $t_0$, the dynamics of the system can be obtained by integrating the ODE system. At time $t_1>t_0$ we obtain
\begin{equation}
u(t_1)=u(t_0)+\int_{t_0}^{t_1}f\bigl(u(t),t;\theta\bigr)\,\mathrm{d}t\,.
\label{eq:integrate}
\end{equation}
This representation clearly demonstrates that the dynamics of the system is entirely encoded in the function $f$, which assigns to each state $u(t)$ and time $t$ the rate of change of this state. The structure of the function $f$ is often dictated to us from physics, and the values of the parameters can be obtained from domain knowledge.\par
Moreover, given an actual numerical implementation of the function $f$ there are several numerical methods, such as Runge-Kutta methods, to integrate ODE systems. Only together with an integration methods will ODEs yield a computational model (\Cref{eq:comp_model}) for predicting future states. 
\end{enumerate}
\paragraph*{Partial differential equations (PDEs)}
A PDE is an equation for a function which depends on more than one independent variable. The equation involves the independent variables, the function, and partial derivatives of the function, with respect to the independent variables. PDEs are ubiquitous in mathematical physics and foundational in several fields, such as acoustics, elasticity, electrodynamics, fluid dynamics, thermodynamics, general relativity, and quantum mechanics. The independent variables are often {\em space-time coordinates}, like $(x,y,z,t)$.\par
As a simple example, we consider a twice differentiable scalar function $u$, which depends on the spatial coordinates $(x,y,z)$, and the PDE
\begin{equation}\label{eq:laplace_eq}
    \frac{\partial^2u(x,y,z)}{\partial x^2}+
    \frac{\partial^2u(x,y,z)}{\partial y^2}+
    \frac{\partial^2u(x,y,z)}{\partial z^2}=0\,.
\end{equation}
This is the {\em Laplace equation} in three dimensions. For example, if $u$ denotes the scalar electric potential, \eqref{eq:laplace_eq} is the governing equation in electrostatics, for domains that are free of electrical charges.

To obtain a computation model (\Cref{eq:comp_model}) for predicting the state of the system over time the PDE will need to be solved either analytically or numerically. Here the finite element method (FEM) is a popular choice, but many other methods exist.

\subsection{Data-driven modeling} 
An alternative path for developing a model is data-centric. Given data in form of observations, a model is developed to be consistent with the observations, for example, reproducing the data as accurately as possible. There are many different data-driven approaches. Unlike the scientific models, which are chosen based on deductive reasoning, data-driven models are chosen based on their statistical and computational properties and their match to the requirements of the modeling problem at hand.

\paragraph*{Data-driven calibration} 
An automotive {\em Electronic Control Unit} (ECU) contains a large number (up to several ten-thousands) of adjustable parameters. The ECU can be seen as a computational model, that processes its inputs such as  requested steering angle and requested acceleration, as well as various sensor measurements, to compute actionable quantities such as the amount of fuel to be injected into the engine. The process of fine tuning the  parameters involved in this computation to ensure optimal performance and efficiency is known as {\em calibration}. For the calibration process, a large amount of data is collected from vehicles that are instrumented with dedicated hardware and software tools. A typical calibration task for a combustion engine could be the optimization of parameters such as injection-, ignition-, or valve-timing with respect to engine outputs such as power, torque, or fuel consumption. 
These parameters are typically adjusted by specialists, in a process of trial-and-error, until the desired behavior is reached.

\paragraph*{Machine learning}
Machine learning presents an approach for learning model parameters from data \citep{bishop2006pattern}. While non-parametric approaches exist, a machine learning model often consists of a parameterized function $u(\cdot\,;\theta)$ with parameters $\theta$, that can predict a response $y$ from inputs $x$. Different parameter settings correspond to different 
functional relationships between the predictions $\hat{y}=u(x;\theta)$ and the inputs. The quality of a prediction, i.e.~how closely a prediction $\hat{y}$ resembles a desired output $y$, can be measured in a {\em loss function} $l(x, y, \theta)$. Given a data set $\mathcal{D}$ of examples of $x$ and $y$ pairs, the optimal parameter setting is found by minimizing the loss, averaged over the training examples,
\begin{align}
\theta^*= \min_\theta \frac{1}{|\mathcal{D}|}\sum_{x,y \in \mathcal{D}}l(x, y, \theta)\,.
\end{align}

\paragraph*{Probabilistic modeling}
Probabilistic modeling refers to a class of machine learning methods where data points are treated as observations of random variables. Modeling consists of making assumptions about the underlying distributions from which these data points are drawn. The primary aim is to infer the parameters that characterize these distributions from the available data. Once the model is learned, it can be used to predict future observations, evaluate the likelihood of observed data, or provide uncertainty estimates regarding the outcomes.

In probabilistic modeling, the uncertainty inherent in predictions is embraced, allowing for more robust decision-making in many scenarios. There are numerous techniques and models in this category, including Bayesian networks, Gaussian processes, Markov chains, and Hidden Markov Models, among others. Each of these models has its own strengths and applications, depending on the nature of the data and the problem at hand. One model class is particularly useful in some hybrid modeling scenarios -- Gaussian processes. For this reason, they are introduced next.

\paragraph*{Gaussian processes}

Gaussian processes (GPs) define a distribution over functions. They provide a principled, non-parametric methodology to infer underlying patterns in data \citep{williams2006gaussian}. A Gaussian process is defined by its mean function $m(x)$ and its covariance or kernel function $
k(x,x')$. At a high level, the mean function describes the expected value of the process, and the kernel function dictates how data points influence each other based on their separation in the input space.

Formally, a Gaussian process can be represented as:
\begin{equation}\label{eq:gaussian_process}
u(x) \sim \mathrm{GP}(m(x), k(x,x')),
\end{equation}
where $u(x)$ is the output of the GP for input $x$, $m(x)$ is the mean function, and $ k(x,x')$ is the kernel function.

Since GPs provide a distribution over functions, they can capture an infinite number of possible explanations for the observed data. Any finite set of these observations can be viewed as being drawn from some multivariate Gaussian distribution defined by the mean and kernel functions. This is particularly powerful as it not only provides a prediction for unseen data but also an associated uncertainty, which can be crucial for decision-making in uncertain environments.

Kernel functions play an integral role in shaping the GP, with the choice of kernel determining the nature of functions the GP can represent. For instance, the Radial Basis Function (RBF) kernel assumes that points closer in input space are more correlated, leading to smooth function approximations. On the other hand, periodic kernels can capture cyclical patterns in the data.

Training a GP typically involves maximizing the likelihood of the observed data under the GP prior, leading to the optimization of kernel hyperparameters. Once trained, predictions with GPs involve conditioning the GP on the observed data to infer values (and uncertainties) at unseen input points.

However, one should note that while GPs offer many advantages, including providing uncertainty estimates and flexibility in modeling, they can become computationally expensive with large data sets. But recent advancements and approximations, like inducing points or sparse GPs, allow for more scalable implementations, making Gaussian Processes a versatile tool for machine learning and hybrid modeling at scale.

\paragraph*{Neural networks}

Neural networks are computational models consisting of interconnected nodes, or ``neurons" (this terminology is borrowed from how the brain processes information), organized into layers: input, hidden, and output layers \citep{goodfellow2016deep}. The connections between neurons has an associated weight, which is adjusted during training to minimize the difference between the predicted and actual output. Each layer of a neural network can be represented as $\sigma(Wx + b)$, 
where $W$ is a matrix of weights, $x$ is the input vector from the previous layer, $b$ is the bias vector, and $\sigma$ represents an activation function, such as the sigmoid or ReLU (Rectified Linear Unit), which is applied element-wise.

The power of neural networks lies in their capacity to approximate complex, non-linear functions. By stacking multiple layers and using non-linear activation functions, neural networks can capture intricate patterns and relationships in data. The training process involves iteratively adjusting the weights using optimization algorithms like gradient descent to reduce the error between the network's predictions and the ground truth.

Deep learning, a sub-field of machine learning, refers to neural networks with many layers, enabling the capture of even more complex representations. For instance, convolutional neural networks (CNNs) are adept at processing image data, while recurrent neural networks (RNNs) excel in handling sequential data.

However, while neural networks have achieved remarkable success in various applications, they come with challenges. Overfitting can occur when a network becomes too complex, and is trained on too little data. Instead of learning the underlying data distribution, it learns to memorize its training data, including the noise. For this reason, training deep networks requires vast amounts of data and computational resources. 

\paragraph*{Regularization of machine learning methods} Regularization techniques serve as foundational tools in machine learning, designed to prevent models from overfitting to their training data. By introducing a penalty to the model's complexity, regularization ensures that models remain generalizable to unseen data~\citep{bishop2006pattern}. $L_1$ (Lasso) and $L_2$ (Ridge) regularization, which penalize the magnitude of model parameters, can be viewed as implicit modeling methods. They don't dictate the model's structure directly but influence it by penalizing certain parameter configurations. In neural networks, techniques like dropout, which randomly deactivates certain neurons during training, aid in enhancing generalization. Other methods such as early stopping and batch normalization, which normalizes neuron activations, further contribute to model robustness. While regularization provides a shield against overfitting, it introduces the challenge of selecting the right regularization strength, necessitating meticulous tuning and validation.

\subsection{Explicit versus implicit models} 
In \Cref{sec:model} we have introduced computational models, and so far avoided the distinction between explicit models, which directly provide computational representations like \Cref{eq:comp_model}, and implicit models, which on their own are not enough to obtain a computational model. While an explicit model prescribes a direct mapping from input x to output y implicit models often require a solver or an optimization procedure to result in a computational model akin to \Cref{eq:comp_model}. Regularization is a fitting example of this distinction. While it introduces constraints or penalties to the learning process, it doesn't directly specify the functional form of the model. Instead, the model emerges as a result of an optimization process that balances fitting the data with the imposed regularization constraints.

Similarly, differential equations provide the dynamics or laws governing a system but don't directly offer a computational model for predicting states. Only when combined with a solver, often numerical, do they yield a method to predict the state at subsequent time points. Partial differential equations (PDEs), such as Maxwell's equations, also epitomize this concept. While they describe the fundamental relationships between electric and magnetic fields, a computational model that predicts field values at specific spatial and temporal points necessitates the application of a solver. The allure of implicit models lies in their ability to capture complex behaviors and constraints. However, they also demand a deeper understanding and careful selection of solvers or optimization techniques to ensure accurate and meaningful predictions.

\subsection{Model composition} 
A computational model, as defined in \Cref{sec:model} can itself be a composition of multiple sub-models. The generic function $u$ that we have used so far can be composed of other functions representing the sub-models in various ways. The sub-models can be implicit or explicit and can be data-driven or first-principles based. The contribution of this paper is to present different design patterns for composing data-driven and first-principle based models.

\paragraph*{Model composition in machine learning} An example of model composition is deep kernel learning \citep{wilson2016deep}. In deep kernel learning, the kernel function of a GP is parameterized using a deep neural network. This means that instead of using a traditional kernel function like the Radial Basis Function (RBF) or Matérn, the kernel is defined by the outputs of a neural network. Formally, given two input vectors 
$x$ and $x'$, the kernel function 
 can be represented as $k_{\theta_k}(f_{\theta_f}(x),f_{\theta_f}(x')) $, where $f_{\theta_f}$ is the neural network with parameters $\theta_f$, and $k_{\theta_k}$, is a base kernel with parameters $\theta_k$.

This composition allows the model to learn intricate patterns and relationships in the data that might not be captured by a standard GP kernel. By mapping the input data into a new representation space using the neural network, the kernel can operate on features that are potentially more informative and better suited to the problem at hand.

Another illustrative example of model composition is the concept of model stacking or stacked generalization \citep{WOLPERT1992241}. Here, individual models, often referred to as base learners, make predictions which are then used as input features for another model, typically called the meta-learner or the stacking model. The meta-learner then makes the final prediction. This composition technique aims to combine the strengths of multiple models, thereby improving generalization performance.

A different perspective on model composition can be found in ensemble methods like bagging \citep{breiman1996bagging} and boosting \citep{schapire1990strength}. In bagging, multiple models are trained on different subsets of the data and then averaged (for regression) or voted upon (for classification) to make predictions. Boosting, on the other hand, iteratively trains models by giving more weight to instances that previous models got wrong, aiming to correct mistakes made by earlier learners.


\paragraph*{Model composition of models based on first principles} Another example of model composition can be found in classical electrodynamics. An electromagnetic field is defined as a four-tuple of space- and time-dependent vector fields $(\vec{E},\vec{D},\vec{H},\vec{B})$, the {\em electric field} $\vec{E}$, the {\em electric displacement} $\vec{D}$, the {\em magnetic field} $\vec{H}$, and the {\em magnetic flux density} $\vec{B}$.
Electromagnetic fields are governed by Maxwell's equations, a set of four PDEs. Two of the equations are dynamic equations, since they contain time derivatives. We collect them in a sub-model $U_1$,
\begin{equation}\label{eq:maxwell_ode}
    \frac{\partial}{\partial t}\begin{pmatrix}\vec{D}\\\vec{B}\end{pmatrix}=\begin{pmatrix} 0&+\curl\\-\curl&0\end{pmatrix}\begin{pmatrix}\vec{E}\\\vec{H}\end{pmatrix}-\begin{pmatrix}\vec{\jmath}\\0\end{pmatrix}\,,
\end{equation}
with the {\em electric current density} $\vec{\jmath}$.
 The first equation in \eqref{eq:maxwell_ode} is Amp\`ere's law, the second Faraday's law, respectively. The remaining two equations have the form of PDE constraints. We collect them in the sub-model $U_2$,
\begin{equation}\label{eq:maxwell_constraint}
    \begin{pmatrix}0\\0\end{pmatrix}=\begin{pmatrix} \Div&0\\0&\Div\end{pmatrix}\begin{pmatrix}\vec{D}\\\vec{B}\end{pmatrix}-\begin{pmatrix}\rho\\0\end{pmatrix}\,,
\end{equation}
with the {\em electric charge density} $\rho$.
These are the electric and magnetic Gauss' laws, respectively. Maxwell's equations $(U_1,U_2)$ need to be complemented by constitutive relations that encode the material properties. For simple media at rest, the additional sub-model $U_3$  takes the algebraic form 
\begin{equation}\label{eq:maxwell_material}
    \begin{pmatrix}\vec{D}\\\vec{B}\end{pmatrix}=
    \begin{pmatrix} \boldsymbol{\varepsilon}&0\\0&\boldsymbol{\mu}\end{pmatrix}
    \begin{pmatrix}\vec{E}\\\vec{H}\end{pmatrix}\,,
\end{equation}
with the dielectric tensor $\boldsymbol{\varepsilon}$ and the permeability tensor $\boldsymbol{\mu}$. All three sub-models can be written in implicit form $U_i(\vec{E},\vec{D},\vec{H},\vec{B})=0$, $i=1,2,3$, and aggregate to the composed model $U=(U_1,U_2,U_3)$, which yields a predictive model of electrodynamics.
\section{Hybrid modeling design patterns}

Hybrid modeling is diverse with applications ranging from molecular modeling in drug discovery, over various simulation tasks in climate science or various engineering disciplines, to modeling sensor data for virtual sensing.  
Solutions for individual use cases are usually application-specific.
New hybrid modeling challenges often seem so unique that interdisciplinary teams come together to develop a custom solution from scratch. While this leads to progress in individual disciplines, solutions are often not accessible to other application domains.

To make progress in \gls{hym} research, it is necessary to abstract recurring modeling challenges and to distill useful solution approaches that generalize across applications. The goal of this paper is to introduce \gls{hym} {\em design patterns} that formalize these solution approaches at an abstraction level beyond individual applications.
We adopt the following definition of design pattern.
\begin{definition}
\label{def:dp}
A hybrid modeling design pattern is a reusable blue-print for a building block of a general solution to recurring hybrid modeling challenges.
\end{definition}
Per our definition, a design pattern should address {\em recurring} challenges beyond individual application domains. For this reason, the solution approach encoded in the design pattern should be {\em general}, meaning that application-specific aspects are abstracted away.  Further, the hybrid modeling design patterns are modular and solving a modeling challenge will typically involve the composition of multiple design patterns. Finally, a design pattern is a blue-print rather than an implementation; blue-prints are {\em reusable} and useful for developing a solution and guiding its implementation.

In this section, we discuss the motivation behind working at this level of abstraction and list properties of useful design patterns. We then introduce the block diagram notation we propose to communicate the design patterns. Finally, we provide some guidance on how the design patterns can be used for new \gls{hym} use cases as well as meta-level research.

\subsection{The block diagram notation for hybrid modeling design patterns}
\label{sec:bd}
We propose a simple block diagram notation for working with the hybrid modeling design patterns. 
The general question in recurring hybrid modeling challenges is typically how to best combine the available domain knowledge with the available data. The data is processed by a data-driven model, which we denote by $D$, while the chosen first-principles-based model is denoted by $P$.
Both models $D$ and $P$ are computational blocks, which receive inputs and perform computations to produce an output. For example, a data-driven model component will receive observations as an input which it will process to either produce a prediction, a lower dimensional representation of the input, or another quantity that is needed for the modeling challenge at hand. The inputs to $P$ will depend on the type of domain knowledge available. In the case of a differential equation for example, the inputs might consist of the initial conditions and the time interval over which the dynamics are to be integrated. The desired output could be the simulated dynamics, or the final state.

In the block diagram notation, a computational block (typically $P$ or $D$) is represented by a square. Directed arrows indicate the flow of information. For example, a directed arrow between two blocks indicates that the output (i.e.~the result of the computation) of the first block, is used as one of the inputs to the second block. A computational block can have multiple incoming arrows, meaning that its inputs come from various sources, and it can have multiple outgoing arrows, meaning that its computational results are further processed in different ways.

 \begin{figure}
 \centering
     \begin{subfigure}[b]{0.18\textwidth}
          \centering
          \resizebox{\linewidth}{!}{
          \begin{tikzpicture}
          \node[font=\LARGE] (pd) at (1.5,1) {$\mathcal{P}$ or $\mathcal{D}$};
          \draw (0,0) rectangle (3,2);
          \end{tikzpicture}
          }  
          \caption{blocks}
          \label{fig:block}
     \end{subfigure}
     \hspace{1.5em}
     \begin{subfigure}[b]{0.18\textwidth}
          \centering
          \resizebox{\linewidth}{!}{
           \begin{tikzpicture}
          \draw[-{Stealth[length=1.5mm, width=1mm]}] (0,0.25) -- (1,0.25);
          \draw[->, white] (0.2,0) -- (0.8,0);
          \end{tikzpicture}
          }  
          \caption{arrows}
          \label{fig:arrow}
     \end{subfigure}
     \hspace{1.5em}
     \begin{subfigure}[b]{0.45\textwidth}
          \centering
          \resizebox{\linewidth}{!}{
          \begin{tikzpicture}
          \draw (0,0) rectangle (3,2);
          \node[font=\LARGE] (b1) at (1.5,1) {$\mathcal{B}_1$};
          \draw (5,0) rectangle (8,2);
          \node[font=\LARGE] (b2) at (6.5,1) {$\mathcal{B}_2$};
          \draw[-{Stealth[length=3mm, width=2mm]}] (3,1) -- (5,1);
\end{tikzpicture}
          }  
          \caption{block diagram}
          \label{fig:simple_bd}
     \end{subfigure}
\caption{\textbf{A block diagram for a hybrid modeling design pattern} consists of computational blocks (\Cref{fig:block}), indicating model components that involve computation, and arrows (\Cref{fig:arrow}) indicating the flow of data and intermediate computational results. For example, the arrow in \Cref{fig:simple_bd} indicates that the result of the block $\mathcal{B}_1$ is fed as an input into the computational block $\mathcal{B}_2$.}
\label{fig:block_diagrams}
 \end{figure}
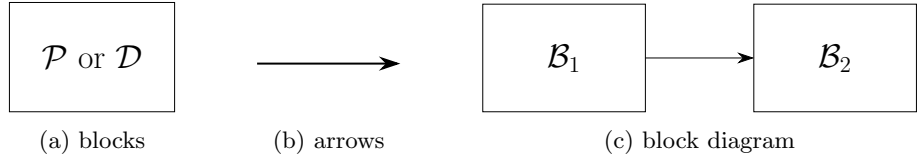

In summary, a block diagram for describing a design pattern consists of rectangular boxes representing computational blocks and of directed arrows, which indicate the flow of inputs and outputs between the boxes. Actual examples of design patterns will be presented in \Cref{sec:dp_examples}.

\subsection{Properties of useful design patterns}
Before diving into the specific design patterns introduced in Section 4 and utilizing the block diagram notation to generate patterns that satisfy Definition 1, it is crucial to discuss the properties that make a design pattern useful. Some of these properties are essential and have already been explicitly stated in our definition of hybrid modeling design patterns.

\paragraph*{Design pattern versus architecture}
We prefer the term ``design pattern'' over ``architecture'' because, in a specific model architecture, several design patterns might be combined or nested. Additionally, we emphasize that the design patterns were collected by analyzing actual applications. Since there is no comprehensive theory of hybrid modeling from which these patterns could be derived, our collection is not exhaustive and is intended to grow as new design patterns are developed or gain importance.

\paragraph*{Abstract and general}An essential step in creating design patterns is abstracting useful concepts that are applicable across various applications and formulating them in a way that makes them easily applicable in a general reusable context. A good design pattern is not a finished design, but rather a blueprint that can be adapted to specific problems.

Design patterns should be abstract and general rather than application-specific, allowing them to be applied across a wide range of problems. This flexibility enables researchers and practitioners to adapt and customize the design pattern for their specific needs, promoting innovation and problem-solving in diverse fields.

\paragraph*{Broad applicability}
A useful design pattern should have the potential to address various challenges and applications, enabling researchers and practitioners to benefit from its adoption. By offering solutions that can be adapted to different contexts, a design pattern with broad applicability can contribute to the development and improvement of numerous models, fostering progress across multiple domains.

\paragraph*{Modularity and composability}
Design patterns should be modular, allowing for easy integration with other patterns, and promoting composability for constructing more complex models. This property enables the combination of multiple design patterns, leading to the creation of more sophisticated and powerful hybrid models that can tackle complex challenges.

\paragraph*{Tractability and ease of communication}
A good design pattern should be tractable, facilitating implementation, and easy to communicate, promoting understanding and collaboration among researchers and practitioners. Clear and understandable design patterns encourage adoption and facilitate the sharing of ideas, contributing to the overall growth and development of hybrid modeling methodologies.

\paragraph*{Clear interface between physics-based and data-driven components}
An effective design pattern should provide a clear interface between the physics-based and data-driven components, enabling seamless integration and interaction between the two modeling paradigms. By defining how these two aspects interact, a design pattern can help create a cohesive and well-structured model that effectively leverages the strengths of both approaches.

\section{Examples of design patterns}
\label{sec:dp_examples}
We now delve into the key design patterns for hybrid modeling. 
There will be two types of patterns, base patterns and composition patterns. 
The base patterns establish systematic approaches for combining a first-principles-based model $P$ with a data-driven model $D$, capitalizing on the strengths of both modeling techniques. In \Cref{sec:basepatterns}, each of the base design patterns is described in detail, elucidating the principles and methodologies underlying their application. Furthermore, we provide illustrative examples to enhance comprehension and demonstrate the practical utility of these design patterns in various scenarios. In \Cref{sec:compatterns}, we present patterns for the composition of base patterns. These composition patterns facilitate building more elaborate hybrid modeling solutions for complex modeling tasks. 

\subsection{Base patterns for hybrid modeling}
\label{sec:basepatterns}
The base patterns are the basic building blocks for the development of hybrid modeling solutions. Each design pattern takes two computational models, typically a first-principles-based model $P$ and a data-driven model $D$ and combines their computation steps into a hybrid model. The order in which the computation is executed, and the flow of inputs and outputs between computational blocks will differ between the design patterns.

In the following sections, we present a total of four base patterns, with the first three having previously been introduced by \Citet{stosch2014hybrid} within the context of process systems engineering.

\subsubsection{The delta model}
\label{sec:delta}
 The delta model serves as a fundamental design pattern in hybrid modeling, providing an effective method to combine the strengths of both first-principles-based and data-driven models. This design pattern is particularly useful when the first-principles-based model captures the primary underlying physical, chemical, or biological processes but may lack the precision or comprehensiveness required for specific applications. 
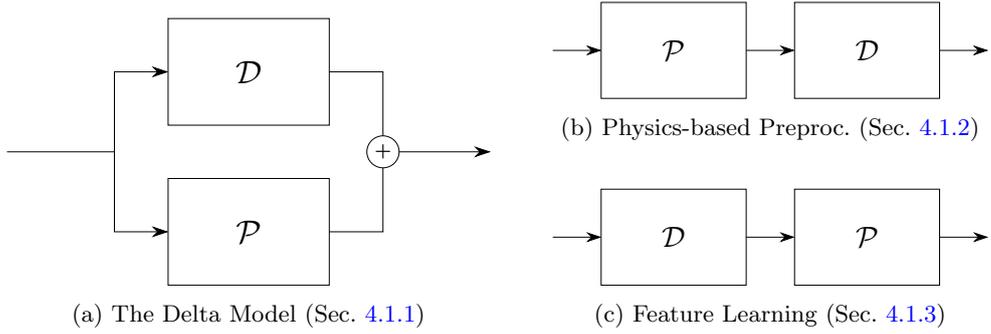
\begin{figure}
    \centering
    \begin{subfigure}[b]{0.5\linewidth}
        \centering
        \resizebox{\linewidth}{!}{
        \begin{tikzpicture}
          \draw (3,0) rectangle (6,2);
          \node[font=\LARGE] (b2) at (4.5,1) {$\mathcal{P}$};
          \draw (3,3) rectangle (6,5);
          \node[font=\LARGE] (b1) at (4.5,4) {$\mathcal{D}$};
          \draw[-{Stealth[length=3mm, width=2mm]}] (0,2.5) -- (2,2.5) -- (2,4) -- (3,4);
          \draw[-{Stealth[length=3mm, width=2mm]}] (2,2.5) -- (2,1) -- (3,1);
          \draw[-{Stealth[length=3mm, width=2mm]}] (6,4) -- (7,4) -- (7,2.5) -- (9,2.5);
          \draw[] (6,1) -- (7,1) -- (7,2.5);
          \draw[fill=white] (7,2.5) circle (0.3cm);
          \node[font=\bfseries] at (7, 2.5) {+};
        \end{tikzpicture}
        }
        \caption{\label{fig:delta_pattern}The Delta Model (\Cref{sec:delta})}
    \end{subfigure}
    \hfill
    \begin{minipage}[b]{0.45\linewidth}
        \begin{subfigure}[b]{\linewidth}
            \centering
            \resizebox{\linewidth}{!}{
            \begin{tikzpicture}
          \draw (3,0) rectangle (6,2);
          \node[font=\LARGE] (b2) at (4.5,1) {$\mathcal{P}$};
          \draw (7,0) rectangle (10,2);
          \node[font=\LARGE] (b1) at (8.5,1) {$\mathcal{D}$};
          \draw[-{Stealth[length=3mm, width=2mm]}] (2,1) -- (3,1);
          \draw[-{Stealth[length=3mm, width=2mm]}] (6,1) -- (7,1);
          \draw[-{Stealth[length=3mm, width=2mm]}] (10,1) -- (11,1);
\end{tikzpicture}
}
            \caption{\label{fig:PBP_pattern}Physics-based Preproc. (\Cref{sec:pbp})}
        \end{subfigure}
        
        \vspace{1em}
        
        \begin{subfigure}[b]{\linewidth}
        \centering
            \resizebox{\linewidth}{!}{
            \begin{tikzpicture}
          \draw (3,0) rectangle (6,2);
          \node[font=\LARGE] (b2) at (4.5,1) {$\mathcal{D}$};
          \draw (7,0) rectangle (10,2);
          \node[font=\LARGE] (b1) at (8.5,1) {$\mathcal{P}$};
          \draw[-{Stealth[length=3mm, width=2mm]}] (2,1) -- (3,1);
          \draw[-{Stealth[length=3mm, width=2mm]}] (6,1) -- (7,1);
          \draw[-{Stealth[length=3mm, width=2mm]}] (10,1) -- (11,1);
\end{tikzpicture}
}
            \caption{\label{fig:feature_pattern}Feature Learning (\Cref{sec:feature_learning})}
        \end{subfigure}
    \end{minipage}
    \caption{Most of the hybrid modeling design patterns can be communicated through block diagrams.}
\end{figure}
 By introducing a data-driven component that accounts for discrepancies or unmodeled phenomena, the delta model can significantly enhance the accuracy and predictive capabilities of the overall hybrid model.

The delta model is formulated by additively combining a first-principles-based model $P$ with a data-driven model $D$, resulting in a hybrid model $H$ as follows:
\begin{equation}
\label{eq:delta_model}
H(x) = D(x) + P(x)\,.
\end{equation}

The block diagram is given in \Cref{fig:delta_pattern}. In the equation, $x$ represents the input variables, and $H(x)$, $P(x)$, and $D(x)$ are the output predictions for the hybrid, first-principles-based, and data-driven models, respectively. The first-principles-based model, $P(x)$, encapsulates the primary knowledge of the underlying processes, while the data-driven model, $D(x)$, is trained to capture the discrepancies between $P(x)$ and the observed data. The data-driven component, therefore, accounts for the unmodeled or inaccurately modeled phenomena, refining the overall predictions made by the hybrid model.

\paragraph*{Typical use cases}

The delta model is applicable in a variety of scenarios, including but not limited to:
\begin{itemize}
    \item Chemical process modeling: \citet{thompson1994modeling} suggest compensating for the inaccuracies of first principle based equations, such as mass and component balances by building a hybrid model which additively combines these simple process models with a neural network.
    \item Ground water modeling in geoscience: \citet{xu2015data} showcase that various data-driven models are effective at correcting the bias of physics-based ground flow models and can in addition produce well calibrated error bars. 
    \item Computational fluid dynamics: Reynold-averaged Navier Stokes (RANS) equation solvers are an important computational tool for modeling turbulent flows. Unfortunately, RANS predictions are often inaccurate due to large discrepancies in the predicted Reynolds stress. \citet{wang2017physic} propose to mitigate these discrepancies with a data-driven correction term.
    \end{itemize}

\paragraph*{Example} 
To study the delta model in action, we consider data from an accelerometer.
The long-term effects can be described by a harmonic oscillator with non-linear dampling, while the short-term effects lack a physical interpretation.
We will study the delta model in comparison to just its physical component $P$ or the data-driven component $D$. We assume, that the underlying dynamics of the system resemble the Van der Pol equation (\Cref{eq:van_der_pol}) and that the short-time behavior can be simulated by a Gaussian process (GP). 

We generate data according to the model

\begin{equation}
y(t) = u_{\text{vdp}}(t) + u_{\text{loc}}(t) + \epsilon,
\end{equation}

where $u_{\text{vdp}}(t)$ are the predictions obtained from the Van der Pol equation, $u_{\text{loc}}(t) \sim \mathrm{GP}(0, k(t,t'))$ are simulated local effects according to a GP with squared exponential kernel with variance $0.2$ and length scale $0.5$ and $\epsilon \sim \mathcal{N}(0, \sigma^2_{n})$ is Gaussian noise with variance $\sigma_n^2=0.05$.

To simulate the Van der Pol equation (\Cref{eq:van_der_pol}), we define the differential $f_{\text{ODE}}:\mathbb{R}^2 \rightarrow \mathbb{R}^2:(s_t, v_t) \rightarrow \bigl(v_t,- s_t+\mu v_t(1 - s_t^2) \bigr)$ in the state-space $h_t=(s_t, v_t)=(u_t, \frac{du_t}{d_t})$, where for ease of readability, we denote a function evaluated at time point $t$ with the subindex $t$, e.g. $u_t \equiv u(t)$.
We use a order 5(4) Runge-Kutta method to simulate $\frac{dh_t}{dt} = f_{\text{ODE}} (h_t; \mu)$ over the time interval $[0,50]$ (at a resolution of $0.1$ units) with $\mu=5$, and initial state $h_0=(1,0)$.

\begin{figure}
\begin{subfigure}[h]{\textwidth}
\includegraphics[width=\textwidth]{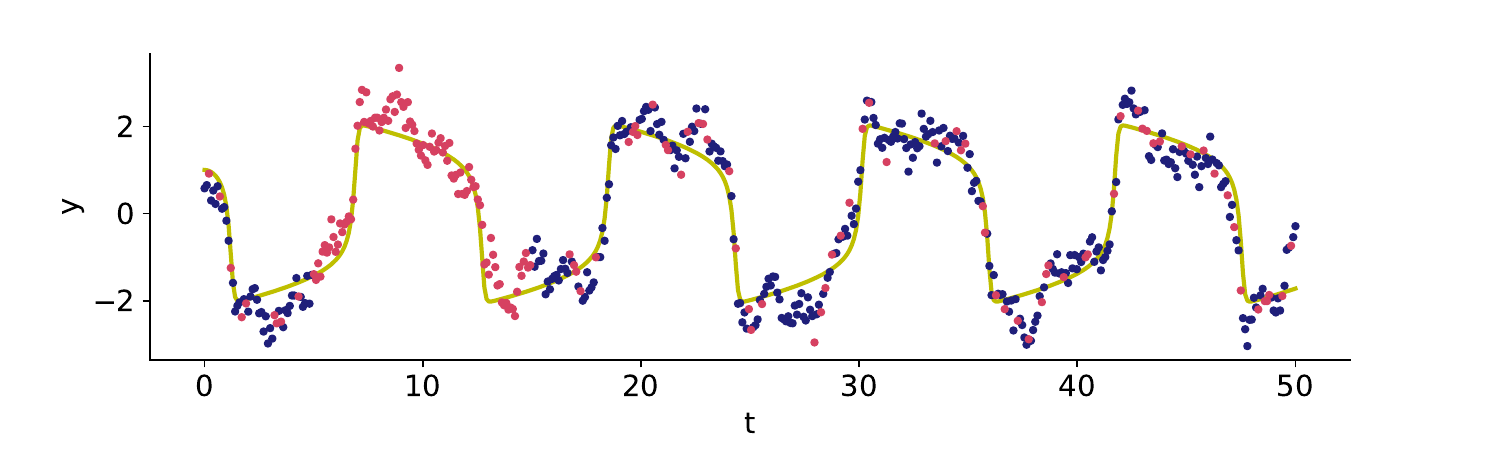}
\caption{Van der Pol oscillator}
\label{fig:van_de_pol}
\end{subfigure}
\begin{subfigure}[h]{\textwidth}
\includegraphics[width=\textwidth]{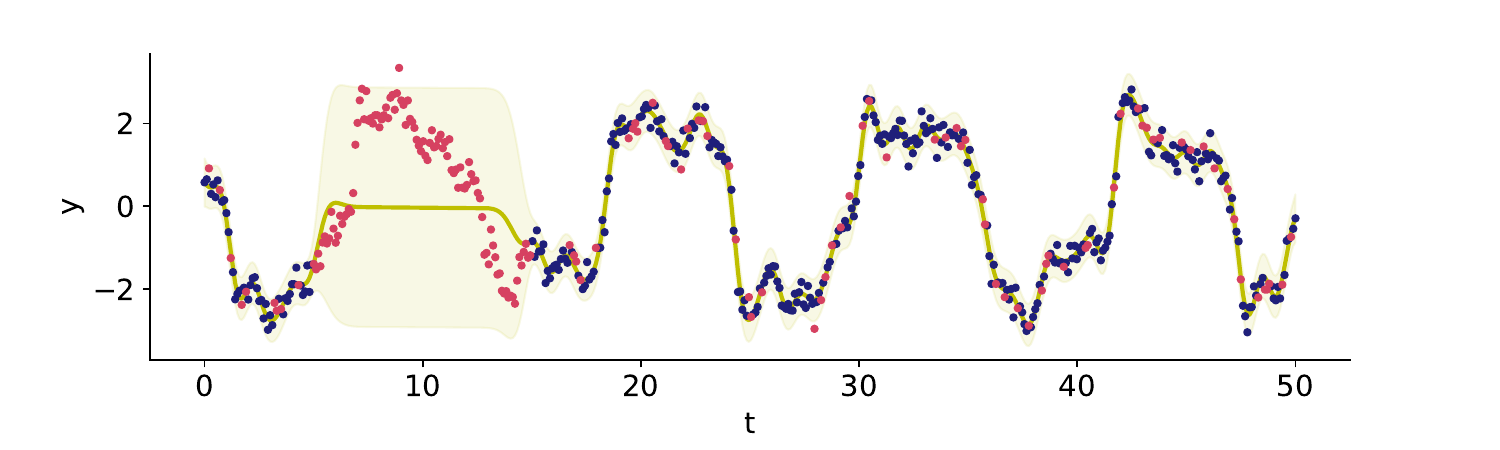}
\caption{Gaussian Process}
\label{fig:gaussian_process}
\end{subfigure}
\begin{subfigure}[h]{\textwidth}
\includegraphics[width=\textwidth]{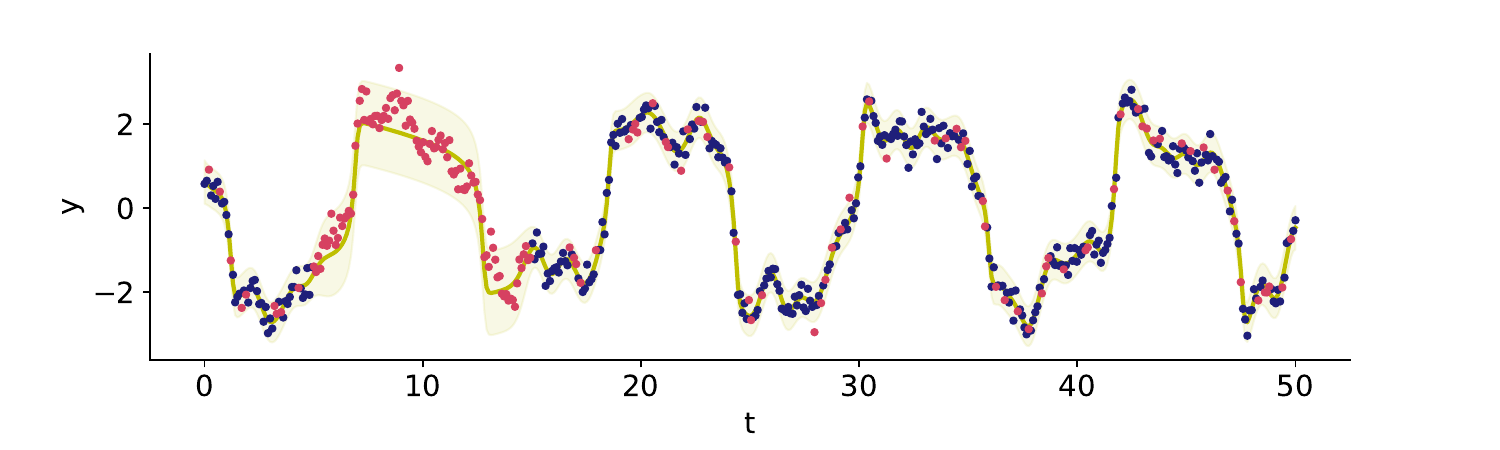}
\caption{Hybrid Model (Delta Pattern)}
\label{fig:hybrid}
\end{subfigure}
\caption{\textbf{Evaluation of different methods on a toy accelerometer set-up.}
From top to bottom: Predictions from a (a) Van der Pol oscillator ($P(t)$), (b) Gaussian Process ($D(t)$) and (c) hybrid model combining both approaches according to the delta model ($H(t) = P(t) + D(t)$).
Training data is shown in blue, test data in red.
We can observe that the Van der Pol oscillator cannot capture the local effects of the data, while the Gaussian Process falls short when training data is scarce.
The hybrid model combines the best of both worlds and performs well under all data scenarios.}
\label{fig:delta_example}
\end{figure}

The generated time series data $\mathcal{D}=(t_k, y_k)_{k=1,\ldots,K}$, where $y_k$ is the measured dynamic response at time $t_k$ is depicted in~\Cref{fig:delta_example}, with training data denoted by blue points and test data denoted by red points. It can be seen that the generated data follows mostly
the Van der Pol equation, which covers the majority of the underlying physical processes, but does not fully account for certain localized phenomena or short-term dynamics.
To make the modeling task more challenging, we further assume that the measurement system had a black-out between 5 and 15 time units during which no training data is available.

The results in the figure provide a qualitative comparison of a pure first principles-based modeling approach based on \Cref{eq:van_der_pol}, fitting a data-based approach (\Cref{eq:gaussian_process}), and a hybrid model using the delta approach.

\Cref{fig:van_de_pol} shows the dynamic response according to the Van der Pol equation. 
While this model accurately captures the long-term behavior of the system, it falls short in capturing the finer details and short-term effects.

The GP predictions are shown in \Cref{fig:gaussian_process}.
When abundant training data is available, the Gaussian Process performs well. 
However, if training data is scarce (between 5 and 15 time units), the predictions fall back to the prior (which is zero) and are accompanied by high uncertainties.

Finally, we combine the Van der Pol oscillator with the Gaussian Process. 
The data-driven model learns the discrepancies between the first-principles-based model’s predictions and the observed data, effectively accounting for unmodeled or inaccurately modeled phenomena.
Results are depicted in \Cref{fig:hybrid} demonstrating that the hybrid model combines the best of both worlds: when training data is available, the Gaussian Process improves the predictions compared to the physics-based model significantly, capturing effects not considered in the Van der Pol equation. 
When training data is limited, the physics-based model takes over, as the Gaussian Process predictions revert to the prior.

Employing the delta model combines the first-principles-based and data-driven
components, resulting in an improved hybrid model.
Our results confirm that this model provides more accurate and reliable predictions by accounting for both the strengths and the limitations of the individual models in different data scenarios.

\paragraph*{Advantages}
The delta model offers several compelling advantages that underscore its utility in hybrid modeling. One of its primary strengths is the facilitation of fast prototyping. With the availability of a first-principles-based model $P$, researchers and practitioners can swiftly initiate their modeling efforts. As more data becomes available or as the need for enhanced precision arises, the data-driven component $D$ can be incrementally introduced, refining the model without necessitating a complete overhaul.

Moreover, the delta model inherently promotes higher accuracy and robustness. While the physical model $P$ provides a foundational understanding, it might occasionally fall short due to assumption mismatches or its inability to encapsulate the stochasticity inherent in many real-world processes. For instance, $P$ might be predicated on idealized assumptions, such as negligible noise levels or presumed linearity, which might not hold true in practical scenarios. The data-driven component $D$ serves as a corrective mechanism in such instances, adeptly learning to account for complex non-linearities, stochastic effects, and other intricate real-world phenomena that the physical model might overlook.

Another salient advantage of the delta model is its data efficiency. Learning the deviations or discrepancies from an existing model $P$ is often more data-efficient than attempting to learn the entire function from scratch solely through $D$. This efficiency is particularly pronounced when training data is sparse. By incorporating the physical model, the delta model introduces a beneficial inductive bias, ensuring that even in low-data regimes, plausible estimates can be generated.

Lastly, the delta model's design inherently supports specialization. In many scenarios, it might be infeasible to obtain training data that spans the entirety of the input domain, perhaps due to safety concerns, prohibitive measurement costs, or other constraints. The delta model elegantly addresses this challenge. For test points that lie outside the domain covered by the training data, the physics-based model $P$ takes precedence, leveraging its capability to extrapolate reliably. Conversely, for inputs that are well-represented in the training data, the data-driven model $D$ offers its specialized insights, ensuring predictions that are both accurate and nuanced.

\subsubsection{Physics-based preprocessing}
\label{sec:pbp}
Physics-based preprocessing is another crucial design pattern in hybrid modeling that leverages domain knowledge to enhance the performance of data-driven models. By incorporating transformations derived from physical laws or other domain-specific knowledge, this design pattern preprocesses the input data before feeding it into a data-driven model. The preprocessing step can introduce useful inductive biases, reduce the dimensionality of the data, and improve the overall efficiency and interpretability of the resulting model.

In the physics-based preprocessing design pattern, a transformation model $P$ is applied to the input variables $x$ before they are fed into a data-driven model $D$. The transformation function incorporates domain knowledge, such as physical laws or constraints, to preprocess the data. The output prediction of the hybrid model $H(x)$ can be expressed as:
\begin{align}
H(x) = D(P(x))\,.
\end{align}

Here, $P(x)$ represents the preprocessed input variables, and $H(x) = D(P(x))$ are the output predictions for the hybrid and data-driven models, respectively. The transformation function, $P(x)$, is designed based on domain knowledge to enhance the data's representation or to simplify the data-driven model's task, leading to improved performance and interpretability. The block diagram for physics-based preprocessing is in \Cref{fig:PBP_pattern}.

\paragraph*{Typical use cases}

Physics-based preprocessing is applicable in various scenarios, including:
\begin{itemize}
\item 
Time-series processing with spectrograms: Time-series data is often preprocessed using short-time Fourier transform (STFT) turning the 1-D time domain signal into a 2-D time-frequency representation. 
Deep learning based methods are more effective in the time-frequency domain for many different applications such as 
time-series anomaly detection \citep{qiu2021neural}, sound classification \citep{hershey2017cnn}, heart disease diagnosis on electrocardiograms \citep{huang2019ecg}
and object classification on radar sensors \citep{patel2019deep}.
\item Fault-detection in mechanical engineering: Rolling-element bearings are an integral component of many machines and bearing fault detection is an important task in mechanical engineering \citep{hamrock1983rolling}. There is a long history of analyzing vibration patterns and acoustic signals for bearing fault detection. For example, peaks in certain spectra are known to be predictive of imminent failure. \citet{sadoughi2019physics} exploit this know-how for physics-based preprocessing of vibration and acoustic data which is then fed into a convolutional neural network (CNN) for bearing fault detection and localization.
\item Demand forecasting: Accurate electricity demand forecasting is an important factor for efficient planning in industry,  healthcare, and urban planning. \citet{bedi2018empirical} combine empirical mode decomposition (EMD) with deep learning. In EMD, the electricity load signals are first decomposed into signals with different time scales, chosen based on domain-knowledge, as well as a residual term. Each of the signal components is then used to train a separate LSTM (Long Short-Term Memory) \cite{hochreiter1997long}. These LSTMs can then be combined to forecast electricity demand.
\end{itemize}
\paragraph*{Example}

\begin{figure}
\begin{subfigure}[h]{0.5 \textwidth}
\begin{center}
\includegraphics[width= 0.7\textwidth]{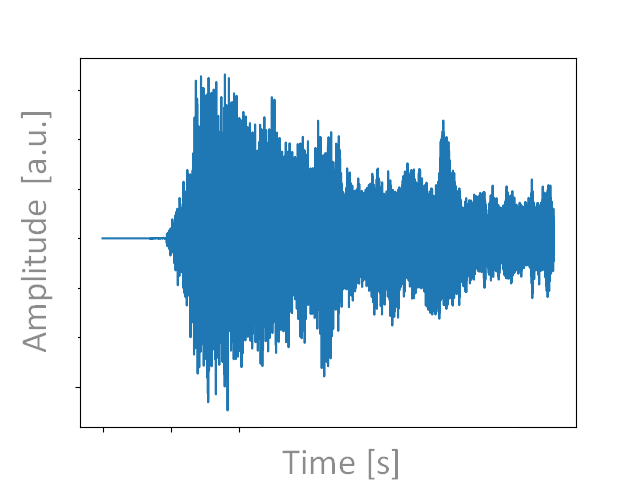}
\end{center}
\caption{Raw data}
\label{fig:audio_raw}
\end{subfigure}
\begin{subfigure}[h]{0.5\textwidth}
\begin{center}
\includegraphics[width= \textwidth]{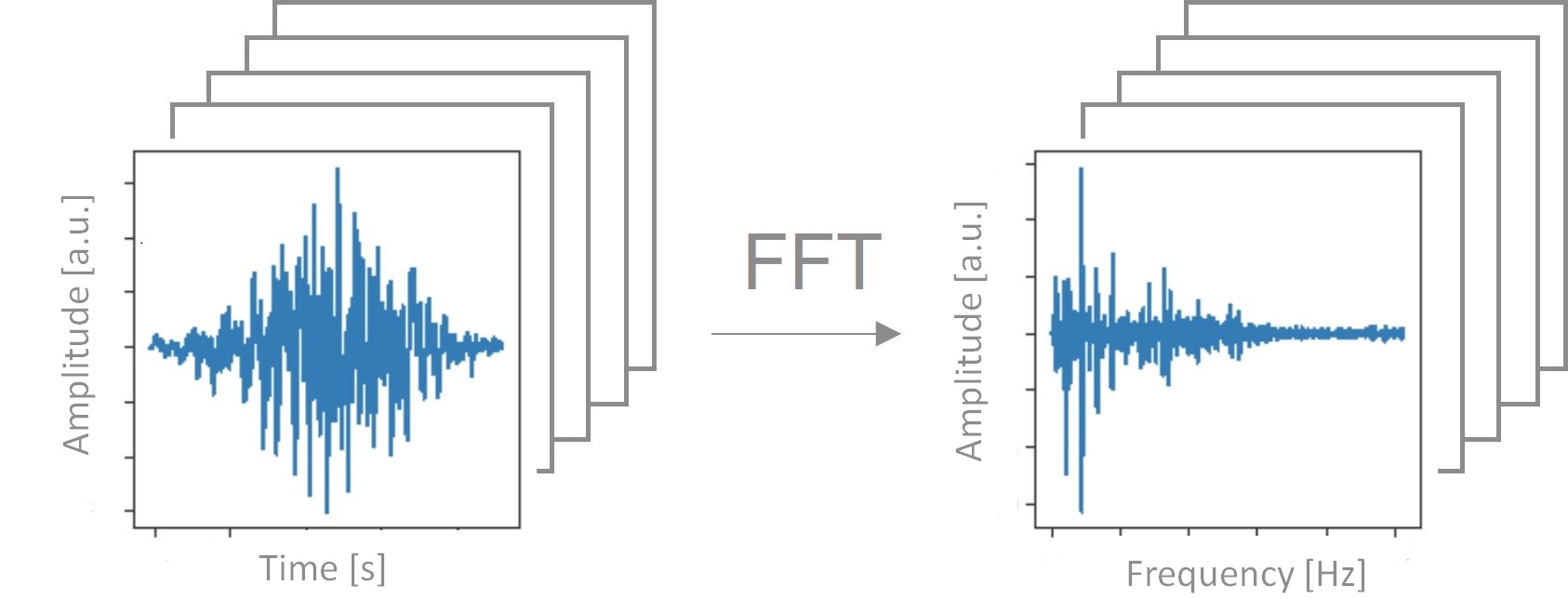}
\end{center}
\caption{Segments}
\label{fig:audio_segments}
\end{subfigure}
\begin{subfigure}[h]{0.5 \textwidth}
\begin{center}
\includegraphics[width= \textwidth]{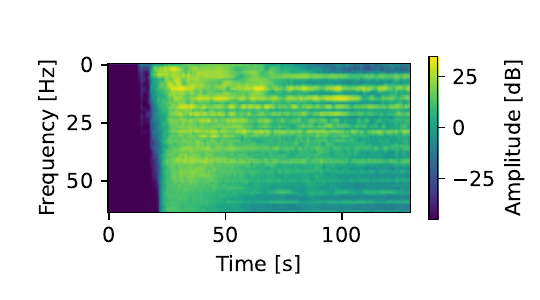}
\end{center}
\caption{Spectrogram}
\label{fig:audio_spectograms}
\end{subfigure}
\begin{subfigure}[h]{.5 \textwidth}
\begin{center}
\includegraphics[width=  \textwidth]{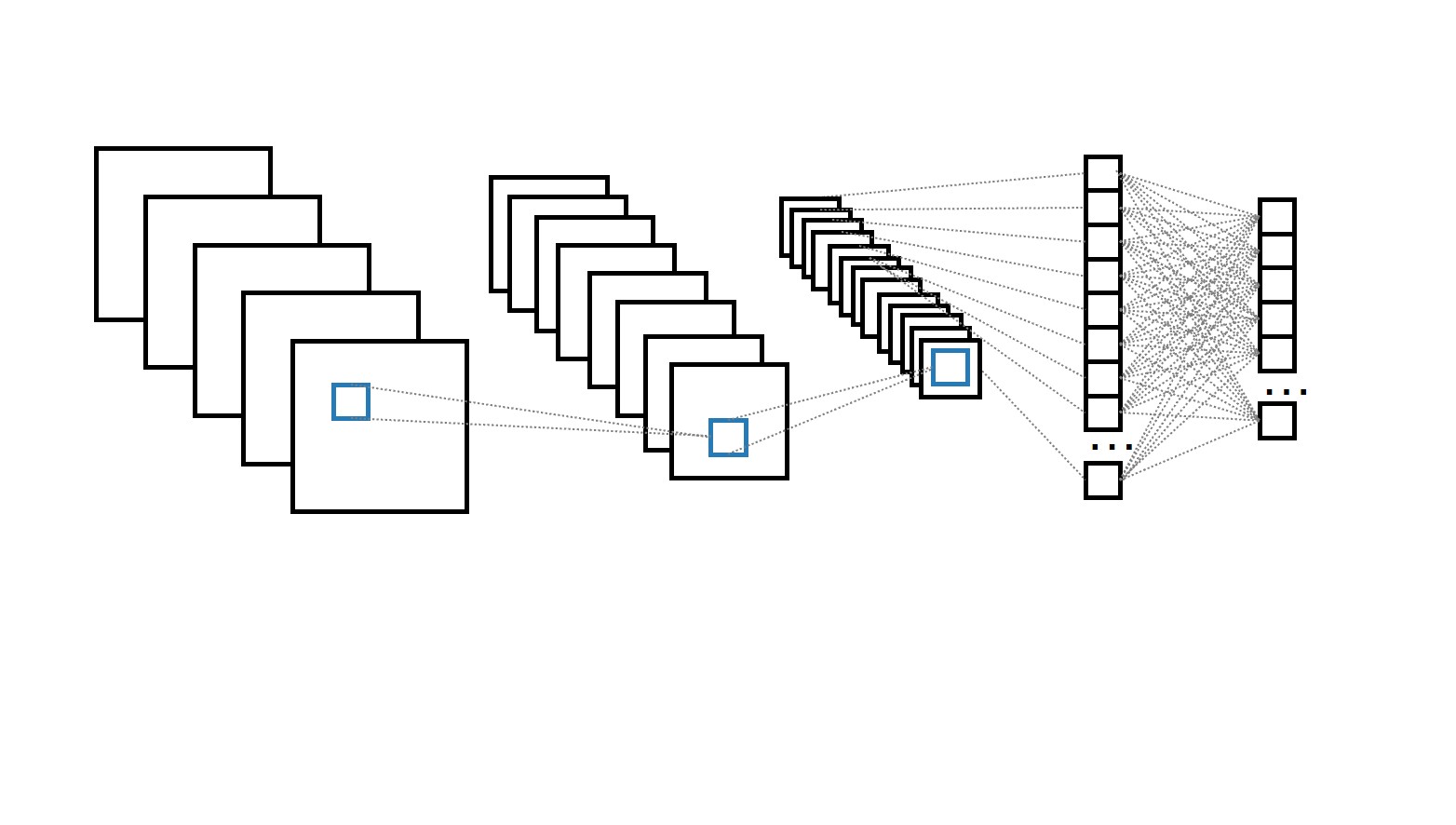}
\end{center}
\caption{Convolutional Neural Network}
\label{fig:audio_cnn}
\end{subfigure}
\caption{Audio Classification with spectrograms}
\label{fig:audio_classification}
\end{figure}
Consider the example of sound classification which is used in many different application fields such as music categorization based on genres, user identification based on voice or bird classification based on audio recordings.

The audio data (see Fig.~\ref{fig:audio_raw}) undergoes an initial transformation into a spectrogram using physics-based preprocessing denoted as P(x). This involves segmenting the audio into overlapping windows of a fixed size (refer to Fig.~\ref{fig:audio_segments}). For each window, a Fourier transform is applied, resulting in a 2-D representation in the time-frequency domain.
Subsequently, each snapshot can be plotted as a Mel spectrogram \citep{RabinerTheoryAA}, where time is represented on the x-axis, frequency on the y-axis, and the amplitude is depicted using colors (see Fig.~\ref{fig:audio_spectograms}).

By obtaining an image representation of the data, we can leverage standard image classification models, denoted as D(P(x)), such as convolutional neural networks (see Fig.~\ref{fig:audio_cnn}). These architectures are designed to respect image structures, incorporating features like translation equivariance and locality. This design choice not only reduces memory requirements but also enhances the model's ability to generalize effectively.

\paragraph*{Advantages}
Physics-based preprocessing in hybrid modeling can improve data efficiency. Using the transformation model $P$ can allow the model to compute features directly, reducing the learning burden on the data-driven model $D$. Especially when $P$ is a type of dimensionality reduction, the lower-dimensional problem can be learned more data efficiently. Note however, that in cases where $P$
does not capture all relevant raw feature information, a purely data-driven model might perform better in data-rich scenarios. This is because $D$ can identify features that outperform human-designed ones, as seen in deep learning methods applied to speech recognition and computer vision.

Similarly, the design pattern also offers resource efficiency. Using pre-computed features in $P$ can simplify the data-based model $D$, potentially removing the need for complex structures like deep neural networks. With features from $P$, simpler algorithms might be adequate for $D$.

Finally, the pattern can increase robustness by avoiding irrelevant feature learning in $D$, that could lead to overfitting or offer an opportunity for adversarial attacks, and it can increase the explainability of the model, by providing a physical interpretation of the features. 

\subsubsection{Feature learning}
\label{sec:feature_learning}
The feature learning design pattern combines data-driven feature learning with downstream physics-based processing. This design pattern comes into play when the first principle based model $P$, for example a controller or a PDE, has some input features that are difficult to measure directly or are difficult to compute precisely from first principles. 

In the feature learning design pattern, a data-driven model $D$ is employed to estimate unmeasurable input variables $v$ based on measurable input variables $x$, $v = D(x)$. These estimated variables are then used as an input for a first-principles-based model $P$ that performs downstream physics-based computations. The output prediction of the hybrid model $H$ can be expressed as:
\begin{align}
\label{eqn:feature}
H(x) = P(x, D(x))
\end{align}

Here, $x$ represents the measurable input variables, and $v = D(x)$ are the estimated unmeasurable input variables produced by the data-driven model. $H(x)$ and $P(x, D(x))$ denote the output predictions for the hybrid and first-principles-based models, respectively. The data-driven model, $D(x)$, is trained to estimate the unmeasurable input variables $v$ using available data, which is then utilized by the first-principles-based model $P(x, D(x))$ for its computations. The block diagram for feature learning is given in \Cref{fig:feature_pattern}. 
In some applications, $D(x)$ will be pre-trained and then combined with $P(x, D(x))$ for hybrid predictions. In other applications, the feature extractor is learned by directly predicting the outputs of the combined hybrid model $H(x) = P(x, D(x))$. This is called end-to-end training.

When $P$ is a physical model, the learned input variables will often have a physical interpretation.
The feature learning design pattern is closely related to the design pattern of physical constraints, which will be discussed in \Cref{sec:constraints}. Since $P$ is used to process the predictions of $D$ we can see $P$ as transforming the outputs of $D$ in a meaningful way, e.g.~to fulfill physical constraints. 

One nuance to consider for the feature learning design pattern is whether $P$ is only used during training, e.g.~to provide a loss or regularization term to guide the data-driven model to make physically plausible predictions, or whether $P$ is also used to make predictions. 
\paragraph*{Typical use cases}
The feature learning design pattern can be applied in various scenarios, including:
\begin{itemize}
\item Electromagnetic field simulations: The optimization of photonic devices requires calculating electromagnetic fields. \citet{DBLP:journals/corr/abs-2203-01248} propose a hybrid approach, where a deep learning model predicts the magnetic near-field distribution. A discrete version of Amp\`ere's law is then used to calculate the electric from the predicted magnetic near field. Eventually, the far field of the outgoing plane wave is computed from the electric near field, by using a near-to-far-field transformation. 
\item Solving PDEs: Deep learning methods for approximating PDE solutions \citep{long2018pde,long2019pde} also exemplify the feature learning design pattern. In these approaches, the model is structured as a PDE, with deep learning techniques employed to learn the differential operators and nonlinear responses of the underlying PDE. This results in models that are capable of capturing complex dynamics while adhering to the physical principles governing the system. 
\item Virtual sensors: Some first-principle-based systems, for example, controllers, require input modalities that are impractical or impossible to measure. For example, a controller for electrical machine torque might require an estimate of rotor temperature \citep{ganchev2011sensorless}. Virtual sensors are data-driven replacements that predict the input modalities that cannot be measured directly but are required for downstream physics-based computations \citep{liu2009virtual}. 
\end{itemize}
\paragraph*{Advantages} 
The feature learning design pattern offers several distinct advantages in hybrid modeling. Firstly, it addresses the challenge of unmeasurable or imprecisely computed input features. By employing a data-driven model $D$ to estimate these features, the pattern effectively bridges the gap between available data and the requirements of a first-principles-based model 
$P$. This not only enhances the accuracy of the hybrid model but also broadens its applicability to scenarios where direct measurements or computations are infeasible. 

This enables virtual sensing, where a predictive model replaces an expensive sensor or enables applications where a required input cannot be measured. In control engineering, this concept is widespread and known as state observer or state estimate.

Secondly, the flexibility of the design pattern allows for both pre-training of $D$ and end-to-end training, catering to different application needs and data availability. This adaptability ensures that the model can be optimized for performance while still benefiting from the strengths of both data-driven and physics-based approaches. When 
$P$ represents a physical model, the learned input variable often carries a meaningful physical interpretation, adding a layer of interpretability to the hybrid model. Furthermore, the integration of 
$P$ ensures that the outputs of 
$D$ are transformed in a manner that aligns with physical constraints or other domain-specific knowledge (this design pattern is described next). This not only enhances the reliability of the model but also ensures that its predictions adhere to known principles, such as the softmax function ensuring outputs that can be interpreted as probabilities. Lastly, the versatility of the pattern allows for 
$P$ to be employed both during training, as a guiding mechanism, and during prediction, ensuring that the model remains grounded in first principles throughout its life cycle.
\subsubsection{Physical constraints}
\label{sec:constraints}
Physical constraints is a hybrid modeling design pattern that incorporates domain knowledge, such as conservation laws, priors, invariances, or statistical independence, to inform the architecture of a data-driven model. The constraints can either affect the structure of the model, the parameters of the model, or its computational results, including both intermediate or final outputs. 

In the design pattern of physical constraints, domain knowledge can be tightly interwoven with the structure or parametrization of a data-driven model $D$. The resulting hybrid model $H$ is formed by incorporating these constraints into the data-driven model, which in its most general form we denote by
\begin{align}
    H(x) = D_P(x).
\end{align}
We choose the notation $D_P$ to indicate that the data-driven model $D$ is informed by physical constraints $P$. The design pattern of physical constraints allows the data-driven model to adhere to the underlying physical principles while still leveraging the benefits of data-driven modeling techniques.

In most of the examples we consider below, the physical constraints are incorporated into model {\em predictions} by first doing the data-driven computations (e.g.~feature extraction with the forward pass of a neural network) and then executing some computational steps derived from first-principles. In this case, the hybrid model can be written as
\begin{align}
    H(x) = P(x, D(x)).
\end{align}

There are many flavors for building hybrid models where a data-driven block $D$ is followed by computation $P$ derived from first-principles. We roughly distinguish three directions: Hard constraints, soft constraints, and feature learning which has already been described. 
In hybrid models with hard constraints, the constraints are implemented in a way such that the predictions of the hybrid model cannot possibly violate the constraints. In contrast, soft constraints, which are often implemented in terms of physics-informed losses for training only approximately guide the predictions to lie within the desired ranges. Feature learning is closely related to the design pattern of hard constraints but has a different motivation. It comes into play, when a model $P$ is missing some input dimensions that cannot be measured and have to be estimated with a data-driven model instead.  

\paragraph{Hard constraints}
\label{sec:hard_constraints}
\paragraph*{Typical use cases}
Hard physical constraints can be applied in various scenarios, such as:
\begin{itemize}
\item Multi-class classification: In multi-class classification, a neural network or another data-driven model $D$ is tasked to produce probabilities over the possible class labels. To ensure that the outputs are in the right range (probabilities are between 0 and 1) and are properly normalized, the last layer is fed through a softmax activation function \citep{bridle1990probabilistic}. This constraint cannot be violated and ensures that the outputs can be interpreted as probabilities. In this example, the constraints affect the output of the model and are part of the model architecture meaning that they take effect both during training and at test time. Also, the softmax implements a {\em hard constraint}; since it is part of the model architecture final predictions cannot violate the desired constraint.
\item Classical mechanics: Hamiltonian neural networks \citep{greydanus2019hamiltonian,toth2019hamiltonian} and Lagrangian neural networks \citep{cranmer2020lagrangian} are another excellent example of this design pattern. In these networks, the model architecture is structured to ensure that the dynamics adhere to conservation laws, such as energy conservation, leading to more accurate and physically meaningful predictions. When modeling the motion of a pendulum, for example, \citet{greydanus2019hamiltonian} use a neural network to directly predict the Hamiltonian of the system. Classical mechanics then determine how to predict the system dynamics, based on the predicted Hamiltonian. Thanks to the Hamiltonian formulation, the structure of the model guarantees that the predicted dynamics conserves energy.
\item Climate modeling: \citet{beucler2019achieving} propose two ways to incorporate linear conservation laws into a neural network for emulating a physical model: By constraining the loss function, or by constraining the architecture itself. Incorporating physical constraints through a loss function is different than modifying model structure: The loss will only guide model outputs to be physically plausible during training. At test time, regularization terms are dropped and while the model might have learned to obey the physical constraints, there are no guarantees that the outputs will be correct. Incorporating physics-based loss terms is therefore an example of {\em soft constraints,} which are discussed next.
\end{itemize}
\paragraph{Soft constraints: surrogates and physics-informed losses}
We have discussed hard constraints, where physical principles are encoded directly into the model structure. An alternative approach for incorporating physical constraints is based on {\em soft constraints.} Here a data-driven model is guided during training to mimic physically plausible behaviour. This is typically achieved by training a surrogate model, i.e.~defining a set of training inputs $\mathcal{X}$ and using training pairs $\{x, P(x) | x \in \mathcal{X}\}$ for training a data-driven model, usually a neural network, to emulate the desired behavior. After training, we will have achieved $D(x) \approx P(x)$ for all $x \in \mathcal{X}$. A related approach for incorporating soft physical constraints is based on physics-based losses. Here the loss function used to  train $D$ will have some term, also called regularization terms, that will encourage $D$ to make physically plausible predictions. These regularization terms can either affect intermediate computation or the final output of the model. In the latter case, the relationship to surrogate modeling becomes clear, as the regularization term will encourage $D(x) \approx P(x)$ for all $x \in \mathcal{X}$. For the design pattern of soft physical constraints, the influence of the physics based model is only explicit during the training phase of model development. At deployment time, the model structure is indistinguishable from a purely data-driven approach. The physical constraints are ``implicitly'' encoded in the parameters of the model.  
\paragraph*{Typical use cases}
Soft physical constraints can be applied in various scenarios, such as:
\begin{itemize}
\item In \cite{liu2019multi}, the authors want to train neural networks to help find solutions of PDEs. For this, they suggest collecting data, where PDEs are solved using the finite element method (FEM). Using this FEM data, the authors train surrogate models that can predict solutions directly. Physical constraints, such as knowledge about the form of the PDE or its boundary values, are incorporated during training via regularization terms. Since high-fidelity solutions are more accurate but more costly to obtain, the authors propose a \textit{multi-fidelity} approach. They train a cheaper low-fidelity surrogate model and a more expensive high-fidelity surrogate model, as well as a  \textit{difference}-NN that can be thought of as a correction term for obtaining a high-fidelity solution from the lower-fidelity one. In this manner, the authors also exploit the delta-model design pattern, in addition to physical constraints.
\item Solving PDEs: Deep learning methods for approximating PDE solutions \citep{long2018pde,long2019pde} also exemplify the physical constraints design pattern. In these approaches, the model is structured as a PDE, with deep learning techniques employed to learn the differential operators and nonlinear responses of the underlying PDE. This results in models that are capable of capturing complex dynamics while adhering to the physical principles governing the system. Physics-Informed Neural Networks (PINNs) \citep{raissi2019physics} demonstrate another application of the physical constraints design pattern. In PINNs, the state of the PDE is parameterized by a neural network, while the structure of the differential operator depends on the specific application, giving rise to the resulting hybrid model. The constraint is included in the loss function. A specialized case of this design pattern is developed by \citet{de2019deep} for advection-diffusion PDEs, which are used for sea surface temperature prediction. A similar approach can be found in \citet{DBLP:journals/corr/abs-2203-01248}, which was also discussed in the context of the feature learning design pattern (\Cref{sec:feature_learning}). A neural network infers the magnetic near-field distribution from the structure of a photonic device. The proposed loss function for training the network contains two additive terms: the usual data-driven loss term and an additional Maxwell loss term, in the spirit of the PINN approach. The Maxwell loss measures the failure of the magnetic field to comply with the vector wave equation. Both loss terms can be balanced by a hyperparameter. The method works most effectively in a regime where more weight is given to data loss. The Maxwell loss can be seen as a regularization, ``\textit{to push the outputted data to be more wavelike}''.
\item Object detection and tracking: Consider the task of learning to detect and track objects in a video. A deep learning approach would typically require labeled examples of input output pairs, such that a neural network (for video typically a CNN) can be trained to predict the outputs given the inputs. \citet{stewart2017label} show that the labeled examples can be replaced by domain knowledge such as physical laws. Instead of using loss functions such as predictive accuracy, they translate physical laws into penalty and regularization terms, yielding loss functions that do not require labels.
\end{itemize}

\paragraph*{Example} 
The design pattern of physical constraints can be used for simulating the electrodynamics of an unknown material. The laws of electrodynamics combine the three sub-models \eqref{eq:maxwell_ode}--\eqref{eq:maxwell_material}. While Max\-well's equations \eqref{eq:maxwell_ode}--\eqref{eq:maxwell_constraint}, i.e., sub-models $U_1$, $U_2$, are accepted as first principles, the constitutive relations \eqref{eq:maxwell_material}, i.e., sub-model $U_3$, is heuristic. Typically an overly simplistic (e.g., polynomial) model is fitted to measurements of material properties. The resulting modeling error compounds when all sub-models are put together.

In \citep{kurz2022hybrid,galetzka2021data} an alternative approach for magnetostatic problems is presented, where the sub-model $U_3$ is discarded altogether. Instead, the authors develop a {\em hybrid solver} that acts directly on the material data to find the best fitting model within all models that are consistent with Maxwell's equations ($U_1$ and $U_2$ in \eqref{eq:maxwell_ode}--\eqref{eq:maxwell_constraint}). In the magnetostatic case, Maxwell's equations reduce to the PDE constraints 

\begin{equation}\label{eq:maxwell_magnetostatic}
\curl\vec{H}-\vec{\jmath}=0\,,\quad\Div\vec{B}=0\,.
\end{equation} 
We denote by $\mathcal{M}$ the space of Maxwell-conforming magnetostatic fields. These are vector fields $z=(\vec{H},\vec{B})$ that exhibit sufficient regularity and are constrained by \eqref{eq:maxwell_magnetostatic}.\par
The measurement data consist of data points $z_i^*=(\vec{H}^*_i,\vec{B}^*_i)$, $i=1,\ldots,N$, that are collected in a set $\widetilde{\mathcal{D}}$. These data are lifted to the space $\mathcal{D}$ of piece-wise constant vector fields $z^*=(\vec{H},\vec{B})$ with respect to a computational grid, such that $\bigl(\vec{H}(x),\vec{B}(x)\bigr)\in\widetilde{\mathcal{D}}$ almost everywhere. Obviously, the data-induced space $\mathcal{D}$ characterizes the magnetic material properties only imperfectly, since it is based on a finite number of measurement points and a spatial discretization by the underlying grid.\par
The solution is formally given by  $\mathcal{S}=\mathcal{M}\cap\mathcal{D}$. These are fields that fulfill Maxwell's equations, while being compatible with the measurement data. However, for a finite number of data points, this set is very likely to be empty. Therefore, we define the solution by the relaxed condition
\begin{equation}\label{eq:relaxed_solution}
    \mathcal{S}=\argmin_{z\in\mathcal{M}}\Bigl(\min_{z^*\in\mathcal{D}}\|z-z^*\|\Bigr)\,,
\end{equation} 
where $\|\cdot\|$ is a suitable norm which serves as loss function. We accept a solution $z$ that conforms to Maxwell's equations, while minimizing the loss function, hence being ``closest'' to the available measurement data.\par
\begin{figure}
    \centering
    \includegraphics[width=0.5\linewidth]{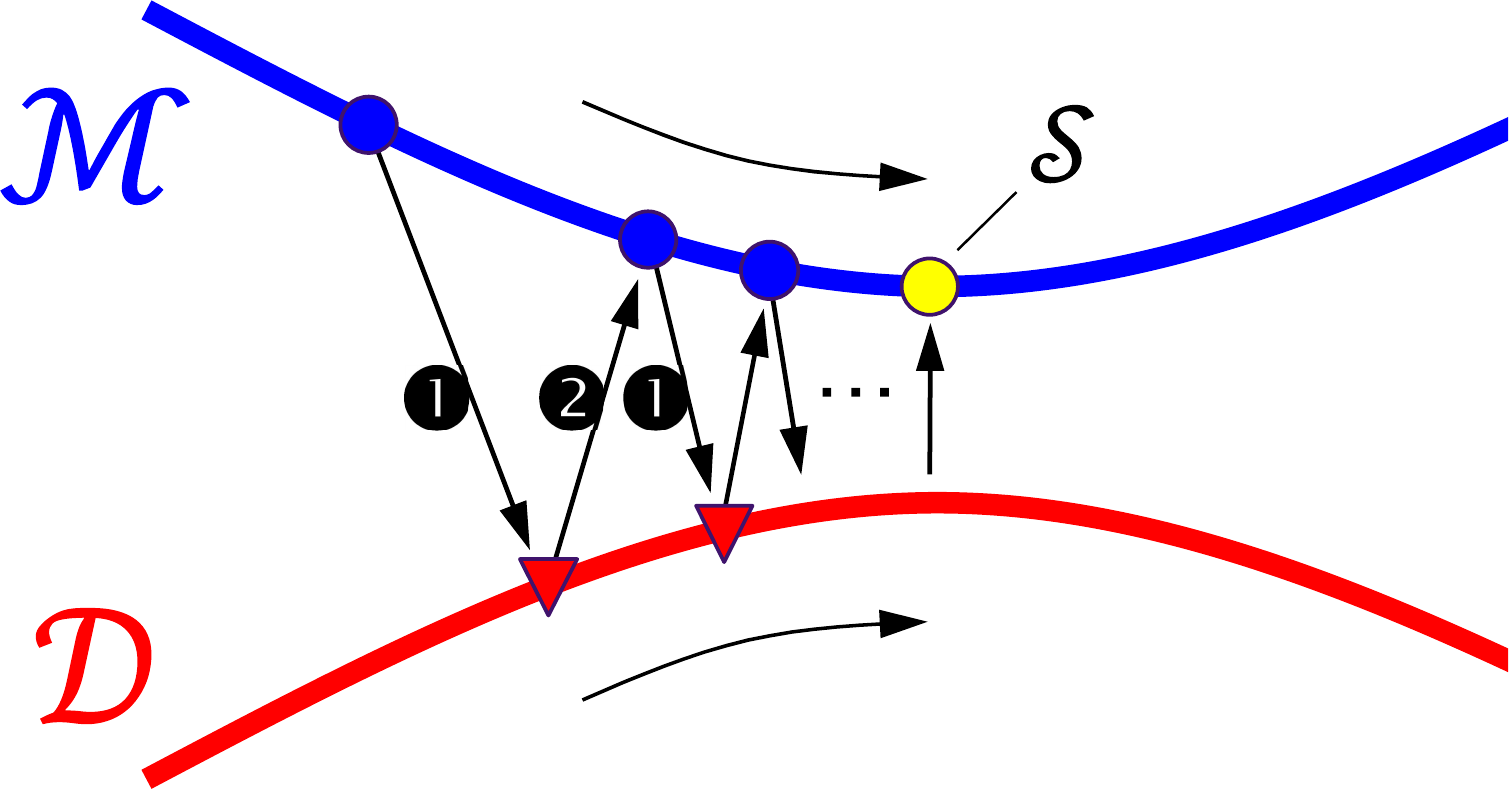}
    \caption{\label{fig:hybrid_solver}Iterative hybrid solver. The fixed point iteration alternates between discrete optimization problems (1) with solutions in the data-induced space (red triangles), and variational problems (2) with solutions in the Maxwell-conforming space (blue circles), the latter being accomplished by a modified finite element solver. [Adapted from \citep{kurz2022hybrid}, Fig.~3]}
\end{figure}
The hybrid solver is organized as a fixed point iteration, see Fig.~\ref{fig:hybrid_solver}. Under convexity assumptions this algorithm converges to the solution of \eqref{eq:relaxed_solution}. Furthermore, it can be shown that the conventional solution is recovered with measurement data sets of increasing size.\par

\paragraph*{Advantages}
Incorporating physical constraints into a hybrid model offers several advantages that enhance both the model's performance and its development process. One of the most pronounced benefits is the combination of higher accuracy with faster design time. The often laborious and data-intensive process of hyper-parameter and architecture optimization can be streamlined by introducing structure through prior knowledge. This not only makes the learning process more feasible but can also lead to more accurate solutions.

Another potential advantage is increased data efficiency. By integrating physical constraints, the complexity (e.g.~dimensionality) of the problem can be reduced, potentially diminishing the volume of required training data. This pre-structuring of the search space accelerates the training of data-based models. Moreover, when $P$ provides a training signal, such as a physically informed self-supervised loss, it can obviate the need for the often expensive labeling process, and instead the training of the data-driven component can benefit from available unlabeled data. 

The design pattern of physical constraints results in hybrid models that benefit from prior knowledge. Priors related to geometry, shapes, invariances, and equivariances, as seen in geometric deep learning \citep{bronstein2017geometric,bronstein2021geometric}, enable the selection of optimal models, bolstering their accuracy and robustness. Furthermore, the explainability of the model is heightened. By grounding the model in physical principles, its topologies become more interpretable, facilitating a clearer understanding of its data-driven components and their interactions with the physical constraints.

\paragraph*{The relationship between physical constraints and feature learning} There are use cases that fit both the physical constraints and the feature learning design pattern, so we describe their relationship here. Unlike hard constraints, soft constraints are only used during the training phase. At deployment time, there is no more computation derived from first principles; instead, the data-driven model has learned to emulate the desired behavior. In contrast, a hard constraint is not removed at deployment time. In \citep{DBLP:journals/corr/abs-2203-01248}, there are hard and soft constraints: a neural network, i.e.~a data-driven model is used to predict the magnetic near field distribution. A soft constraint based on Maxwell's equations, ensures that the predictions adhere with the laws of physics. These predictions are then processed by a computational block $P$ that implements a discrete version of Amp\`ere's law, followed by a near-to-far field transformation. $P$ can be interpreted as imposing a hard constraint since it is guaranteed to produce a prediction of the electric field that is consistent with the magnetic field prediction of $D$. The constraint is used both during training and at test time. In this example, the soft constraint is on an intermediate output of the model, while the hard constraint affects the final output of the model. In general, constraints can either affect intermediate of final computation, or parameter values of the model, or the structure of the model. Note that a hybrid modeling solution, where a computational block $D$ is followed by a hard constraint, i.e.~a constraint that is not removed after training and that affects the final computational output, is consistent with \Cref{eqn:feature} and therefore also fits the feature learning design pattern. In fact, \citep{DBLP:journals/corr/abs-2203-01248} was presented as an example of the feature learning design pattern in \Cref{sec:feature_learning} for that reason.

It is quite common for hybrid modeling solutions to combine multiple design patterns. In the next section, we describe design patterns for pattern composition.

\subsection{Composition patterns for hybrid modeling}
\label{sec:compatterns}
Next, we describe composition patterns. They provide patterns for composing the base patterns from \Cref{sec:basepatterns} into more elaborate hybrid modeling solutions.
\subsubsection{Recurrent composition}
An important design pattern, especially when dealing with sequential data, is recurrent composition.
The recurrence design pattern encompasses a wide range of models involving an internal state that is updated sequentially. This pattern is observed in recurrent neural networks and numerical integration schemes for differential equations. The main principle is to compute the dynamics of a system through a recursive update rule. The computational block $H$ for the update rule can either be data-driven, or based on first principles, or consist of a hybrid computational block that relies on one or more of the design patterns presented above.

The recurrence design pattern features an internal state $s$ which is updated sequentially over time. The state at time $t$ is computed from a previous state:
\begin{align}
    s_t = H(s_{t-1}, \cdots)
\end{align}
The function $H(\cdot)$ can have additional inputs, such as observations from a sequence $x_1$, $x_2$, $\cdots$, $x_T$, the time $t$, and the time difference $\Delta_t$ between $s_{t-1}$ and $s_t$. In control or signal processing applications, there might also be a control input. Whether $H$ is data-driven, physics-based, or hybrid, depends on the use-case. Some typical use cases are described next.

\paragraph*{Typical use cases}
\begin{itemize}
\item Recurrent neural networks in deep learning: \Glspl{rnn} are powerful sequence models. When trained on sequences of observations $x_1$, $x_2$, $\cdots$, $x_T$, they have the capacity to leverage $s_t$ as a hidden state to summarize all the relevant information in the sequence up until time $t$. At each time step the hidden state is updated based on the current observation and the previous hidden state $s_t = H(s_{t-1}, x_t)$.
To obtain a prediction, the hidden state can then be mapped to the desired output. For a vanilla \gls{rnn}, $H(s_{t-1}, x_t)$ will be an affine transformation followed by a non-linearity, but other choices exist, such as \glspl{gru} \citep{chung2014empirical} and \glspl{lstm} \citep{hochreiter1997long}. For most \glspl{rnn}, $H$ is data-driven, meaning that the parameters are learned by fitting to training data \citep{goodfellow2016deep}.
\item Numerical integration: A dynamical system is often described by an ODE as in~\Cref{eq:linear_system_odes}. Some \glspl{ode} allow recovering the system state using analytic solutions but in many interesting cases numerical integration schemes have to be employed to compute the state of the system as a function of time. In a numerical integration scheme, the system state is approximated by $s_t$, which can also be thought of as the intermediate integration results at time $t$. Typically, there is a recursive update rule where $s_t$ is computed based on a previous state $s_{t-1}$ as well as the step size and the vector field $f$. In the backward Euler method for example $s_t = H(s_{t-1}, \Delta_t, t) = s_{t-1} + \Delta_t f\bigl(s_t,t;\theta\bigr)$, 
with $f$ and $\theta$ as defined in \Cref{eq:linear_system_odes}.
\item Neural \glspl{ode}: Neural \glspl{ode} \citep{chen2018neural} are a model class at the intersection of deep learning and differential equations. The vector field $f$ in~\Cref{eq:linear_system_odes} is parameterized by a neural network. The result is a flexible dynamics model whose parameters are fitted in a data-driven way. Neural \glspl{ode} rely heavily on numerical integration: The system has to be integrated to form a prediction, and back-propagation through the \gls{ode} solver can be handled efficiently by numerically integrating an auxiliary (adjoint state) \gls{ode} backward in time \citep{lecun1988theoretical}.
\item State estimation: State estimation is a crucial process in control theory and signal processing that aims to accurately determine the state of a dynamic system based on noisy and potentially incomplete measurements over time~\citep{sarkka2023bayesian}. The relationship between the inputs and the outputs of the dynamical system is often described by \glspl{ode}. In addition to predicting the system state by (numerical) integration of the dynamics, state estimation also entails accounting for the influence of control inputs, and for measurement noise, thereby systematically improving the accuracy of the system state's prediction. One notable example of an algorithm used for state estimation is the K\'alm\'an filter \citep{kalman1960new}, which provides the optimal solution to estimate the state of a linear dynamic system perturbed by Gaussian noise. For state estimation in non-linear systems, variations such as the Extended K\'alm\'an Filter (EKF) or Unscented K\'alm\'an Filter (UKF) are often used \citep{julier2004unscented}.
\end{itemize}

\paragraph*{Example} Modern recurrent neural networks typically assume regular time intervals between observations. A notable exception is the \gls{cru} which can be used to model irregularly sampled time series \citep{schirmer2022modeling}. It assumes a hidden state that evolves according to a linear \gls{sde}. To model a sequence, each measurement is first mapped into a latent space by a neural network. The transformed observation is then treated as an observation of the latent state, which can now be inferred via state estimation, specifically the continuous-discrete formulation of the K\'alm\'an filter \citep{jazwinski2007stochastic}. 

\begin{figure*}[t]
    \centering
    \includegraphics[width=\textwidth]{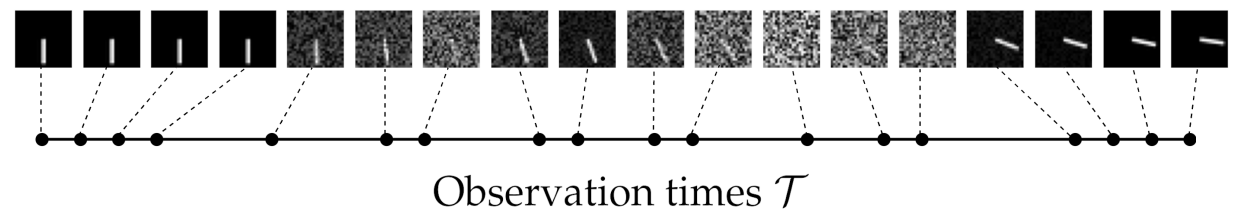}
    \caption{The CRU \citep{schirmer2022modeling} can help infer the pendulum angle from images observed at irregular time intervals.}
    \label{fig:noise}
    \vskip -0.1in
\end{figure*}

The recursive update of the \gls{cru} is a hybrid block, combining a data-driven block $D$, which consists of a neural network and is applied to each measurement $x_t$, and a state estimation block $P$ consisting of the update equations of the continuous-discrete K\'alm\'an Filter. 
\begin{align}
    s_t = H(s_{t-1}, x_t, \Delta_t) = P(s_{t-1}, D(x_t), \Delta_t)
\end{align}
As an illustrative example, consider the problem of predicting the angle of a pendulum from noisy images taken at irregular time intervals (\Cref{fig:noise}). Since some of the images are very noisy, angle prediction will benefit from a model that takes temporal structure into account, such as the \gls{cru}. While the pendulum dynamics are relatively simple and can be described by a second-order \gls{ode}, inferring them from high-dimensional inputs such as images is non-trivial. 
The \gls{cru} can accurately predict the angle, optimally accounting for different sources of noise.

\paragraph*{Advantages} The concept of recurrence is useful in hybrid modeling and machine learning for several reasons. First, recurrent models can learn to recognize patterns across time. For example, they can learn to predict the next word in a sentence based on the context provided by the preceding words. This is possible because the model has a way of remembering the previous context, enabling it to learn how the current state is influenced by the previous states.

Another advantage of this design pattern is parameter sharing. Recurrent models apply the same set of weights to the inputs at each time step. This means that they are making the assumption that the same patterns that are useful to recognize at one point in time will be useful to recognize at other points in time. This significantly reduces the number of parameters in the model, which can help to avoid overfitting and make the model easier to train.

Finally, recurrence provides a natural modeling paradigm to deal with input and output sequences of variable length. For example, you can use an RNN to process a sentence of any length and produce a sentiment score. Traditional methods like feed-forward neural networks cannot handle this variability as they require fixed-size input vectors.

\subsubsection{Hierarchical pattern composition}
The pattern of pattern composition emphasizes the flexibility and composability of hybrid modeling design patterns. In this pattern, the concept is that hybrid models themselves can serve as building blocks for constructing more complex hybrid models. To represent this idea, we introduce the following notation:

Let $H(P, D)$ denote a hybrid model that combines a physics-based model $P$ and a data-driven model $D$. The pattern of pattern composition suggests that $P$ and $D$ themselves can be hybrid models. We can represent this idea by considering two hybrid models, $H_1$ and $H_2$, such that:

\begin{equation}
H(P, D), \quad \text{where} \quad P = H_1(P_1, D_1) \quad \text{and} \quad D = H_2(P_2, D_2).
\end{equation}

This notation conveys that $H_1$ and $H_2$, each being a combination of physics-based and data-driven models, are now being combined to form a new, more complex hybrid model $H$. This pattern highlights the recursive nature of hybrid modeling, where models can be built upon one another in a hierarchical manner, leading to increasingly sophisticated representations of the underlying system.

By applying the pattern of pattern composition, practitioners can create multi-layered hybrid models that address various aspects of the problem at hand, and tackle more complex challenges by leveraging the strengths of multiple modeling paradigms. This approach also allows researchers to explore novel combinations of the design patterns introduced in this paper, potentially leading to new insights and advances in the field of hybrid modeling.
\paragraph*{Typical use cases}
\begin{itemize}
\item Lake Temperature Modeling: \citet{daw2017physics} present a hybrid modeling solution for lake temperature modeling. The goal is to predict temperature from physical quantities that are known to drive lake temperature. The authors assume access to observations and a physics-based simulation of lake temperature $P_1$, which might be inaccurate due to inadequate calibration or missing physics. The physics-based pre-processing design pattern is used to first augment the input variables with the potentially inaccurate but still useful predictions of $P_1$. The original observed features $x$ are concatenated with these physically preprocessed predictions to  $[x, P_1(x)]$, which is then fed into a data-driven model that is further subjected to the design pattern of physical constraints. An additional loss term $P_2$ assures that the predictions fulfill plausible density-depth and density-temperature relations. The combined hybrid model can be written as $H(x) = D_{P_2}([x, P_1(x)])$.
\item ODEs with missing physics: Another example of hierarchical pattern composition is a hybrid neural ODE \citep{yin2021augmenting} where the vector field $f$ of the ODE in \Cref{eq:linear_system_odes} is parameterized by multiple terms which are added according to the delta model design pattern. This can be beneficial when part of the dynamics are explicitly known, while other missing parts are modeled in a data-driven way, typically with a neural network.
Extensions to stochastic dynamical systems also exist \citep{haussmann2021learning}.
\item Dynamics modeling with unknown unknowns: \citet{long2018hybridnet} propose a hybrid model for dynamics modeling with many unknowns. For example, in a fluid dynamics application, it is known that the dynamics are
governed by Navier-Stokes equations, but they cannot be solved without knowledge of the geometry of
the system or access to physical parameters such as viscosity, material density, or external forces. In such a setting the authors suggest employing a learnable PDE solver $H_1$ based on cellular neural networks. This learnable PDE solver can be seen as a hybrid approach: it is a data-driven approach where missing physical parameters are learned from data, but its structure is derived from first-principles and adheres to the underlying PDE. 
To deal with missing inputs, e.g.~with unobserved external perturbations to the inputs, the authors further employ the feature learning design pattern. A data driven model $D$, specifically a convolutional LSTM predicts the missing inputs, which are then fed into $H_1$, resulting in the composed hybrid model $H_2(x) = H_1(D(x))$.
\end{itemize}

\paragraph{Example}

Many time-series algorithms face challenges when attempting to simultaneously capture short- and long-term effects. Data-driven models (denoted as $D$) often excel at providing detailed short-term predictions. However, even small errors in their short-term forecasts can accumulate over time, leading to deteriorated long-term performance.
In contrast, models capable of reliable long-term predictions can often be developed by leveraging physics-based simulations (referred to as $P$).

The work of \cite{ensinger2023combining} addresses this challenge by decomposing predictions into two components: one that accurately predicts long-term behavior and another one that excels at short-term prediction. The long-term predictions are generated by the physics-based model $P$, while the short-term predictions are generated by the data-driven model $D$. To ensure that each model operates within its domain of competence, the authors introduce two hard constraints: 
They apply a low-pass filter ($F_{\text{low}}$) to the predictions of the physics-based model $P$ and a high-pass filter ($F_{\text{high}}$) to the predictions of the data-driven model $D$.
Finally, the two prediction components are combined using the delta pattern resulting in a complementary filtering approach:
\begin{eqnarray}
H(x) = F_{\text{low}}\bigl(P(x)\bigr) +  F_{\text{high}}\bigl(D(x)\bigr).
\end{eqnarray}

The fusion of high and low-frequency information from different signals is a well-established technique in control engineering and signal processing applications. 
An illustrative example can be found in robotics, specifically in tilt estimation\citep{trimpe2010accelerometer}. 
In this context, accelerometer and gyroscope measurements are often recorded simultaneously. 
The gyroscope delivers precise short-term position estimates, but due to integration at each time step, accumulating errors introduce drift in the long-term. 
In contrast, accelerometer-based position estimates are more stable over the long-term but exhibit substantial noise, making them less reliable for short-term predictions.
As a consequence, the position estimate can be significantly improved by combining both signals after applying a high-pass filter to the gyroscope measurements and a low-pass filter to the accelerometer measurements.
 
\paragraph*{Advantages}
Only through composition do the design patterns reach their full potential. While here we have provided three examples, for how design patterns can be composed, the possibilities are endless. While each of the design patterns has their own set of advantages, through composition we can build hybrid models that combine many of these advantages into a single modeling solution.

\section{Conclusion}
In conclusion, this paper has presented a systematic exploration of various design patterns for hybrid modeling, showcasing the potential of combining the strengths of both data-driven and mechanistic models to address complex problems in diverse domains. These design patterns provide a unified framework for understanding and organizing the myriad approaches used in hybrid modeling, and they facilitate the sharing of knowledge and expertise across application domains.

The identification and formalization of these design patterns serve as a valuable resource for researchers and practitioners in the field, allowing them to better understand the underlying principles, common challenges, and potential solutions for hybrid modeling. By providing a higher level of abstraction, these design patterns enable the development of more generalizable and standardized tools and techniques, leading to improved efficiency and reliability of the modeling process.

Furthermore, the use of design patterns can help to identify common limitations and areas for improvement in hybrid modeling, thus guiding future research directions and fostering innovation. As the field of hybrid modeling continues to evolve, we anticipate that the exploration and refinement of these design patterns will play a crucial role in shaping the development of new models, methods, and applications, ultimately contributing to the advancement of our understanding and the solution of real-world problems.

In summary, the design patterns presented in this paper offer a valuable framework for organizing and advancing the field of hybrid modeling. By embracing the principles of abstraction and generalization, researchers and practitioners can better address the unique challenges and complexities of their domains, while also contributing to the broader knowledge and understanding of hybrid modeling as a whole.

\bibliographystyle{plainnat}
\bibliography{hym_refs}

\begin{thebibliography}{59}
\providecommand{\natexlab}[1]{#1}
\providecommand{\url}[1]{\texttt{#1}}
\expandafter\ifx\csname urlstyle\endcsname\relax
  \providecommand{\doi}[1]{doi: #1}\else
  \providecommand{\doi}{doi: \begingroup \urlstyle{rm}\Url}\fi

\bibitem[Bedi and Toshniwal(2018)]{bedi2018empirical}
Jatin Bedi and Durga Toshniwal.
\newblock Empirical mode decomposition based deep learning for electricity demand forecasting.
\newblock \emph{IEEE access}, 6:\penalty0 49144--49156, 2018.

\bibitem[Beucler et~al.(2019)Beucler, Rasp, Pritchard, and Gentine]{beucler2019achieving}
Tom Beucler, Stephan Rasp, Michael Pritchard, and Pierre Gentine.
\newblock Achieving conservation of energy in neural network emulators for climate modeling.
\newblock \emph{arXiv preprint arXiv:1906.06622}, 2019.

\bibitem[Bishop and Nasrabadi(2006)]{bishop2006pattern}
Christopher~M Bishop and Nasser~M Nasrabadi.
\newblock \emph{Pattern recognition and machine learning}, volume~4.
\newblock Springer, 2006.

\bibitem[Breiman(1996)]{breiman1996bagging}
Leo Breiman.
\newblock Bagging predictors.
\newblock \emph{Machine learning}, 24:\penalty0 123--140, 1996.

\bibitem[Bridle(1990)]{bridle1990probabilistic}
John~S Bridle.
\newblock Probabilistic interpretation of feedforward classification network outputs, with relationships to statistical pattern recognition.
\newblock In \emph{Neurocomputing: Algorithms, architectures and applications}, pages 227--236. Springer, 1990.

\bibitem[Bronstein et~al.(2017)Bronstein, Bruna, LeCun, Szlam, and Vandergheynst]{bronstein2017geometric}
Michael~M Bronstein, Joan Bruna, Yann LeCun, Arthur Szlam, and Pierre Vandergheynst.
\newblock Geometric deep learning: going beyond {Euclidean} data.
\newblock \emph{IEEE Signal Processing Magazine}, 34\penalty0 (4):\penalty0 18--42, 2017.

\bibitem[Bronstein et~al.(2021)Bronstein, Bruna, Cohen, and Veli{\v{c}}kovi{\'c}]{bronstein2021geometric}
Michael~M Bronstein, Joan Bruna, Taco Cohen, and Petar Veli{\v{c}}kovi{\'c}.
\newblock Geometric deep learning: Grids, groups, graphs, geodesics, and gauges.
\newblock \emph{arXiv preprint arXiv:2104.13478}, 2021.

\bibitem[Chen et~al.(2022)Chen, Lupoiu, Mao, Huang, Jiang, Lalanne, and Fan]{DBLP:journals/corr/abs-2203-01248}
Mingkun Chen, Robert Lupoiu, Chenkai Mao, Der-Han Huang, Jiaqi Jiang, Philippe Lalanne, and Jonathan~A. Fan.
\newblock Wavey-net: Physics-augmented deep learning for high-speed electromagnetic simulation and optimization.
\newblock \emph{CoRR}, abs/2203.01248, 2022.
\newblock URL \url{https://doi.org/10.48550/arXiv.2203.01248}.

\bibitem[Chen et~al.(2018)Chen, Rubanova, Bettencourt, and Duvenaud]{chen2018neural}
Ricky~TQ Chen, Yulia Rubanova, Jesse Bettencourt, and David~K Duvenaud.
\newblock Neural ordinary differential equations.
\newblock \emph{Advances in neural information processing systems}, 31, 2018.

\bibitem[Chung et~al.(2014)Chung, Gulcehre, Cho, and Bengio]{chung2014empirical}
Junyoung Chung, Caglar Gulcehre, Kyunghyun Cho, and Yoshua Bengio.
\newblock Empirical evaluation of gated recurrent neural networks on sequence modeling.
\newblock \emph{arXiv preprint arXiv:1412.3555}, 2014.

\bibitem[Cranmer et~al.(2020)Cranmer, Greydanus, Hoyer, Battaglia, Spergel, and Ho]{cranmer2020lagrangian}
Miles Cranmer, Sam Greydanus, Stephan Hoyer, Peter Battaglia, David Spergel, and Shirley Ho.
\newblock Lagrangian neural networks.
\newblock \emph{arXiv preprint arXiv:2003.04630}, 2020.

\bibitem[Daw et~al.(2017)Daw, Karpatne, Watkins, Read, and Kumar]{daw2017physics}
Arka Daw, Anuj Karpatne, William Watkins, Jordan Read, and Vipin Kumar.
\newblock Physics-guided neural networks {(PGNN)}: An application in lake temperature modeling.
\newblock \emph{arXiv preprint arXiv:1710.11431}, 2017.

\bibitem[De~B{\'e}zenac et~al.(2019)De~B{\'e}zenac, Pajot, and Gallinari]{de2019deep}
Emmanuel De~B{\'e}zenac, Arthur Pajot, and Patrick Gallinari.
\newblock Deep learning for physical processes: Incorporating prior scientific knowledge.
\newblock \emph{Journal of Statistical Mechanics: Theory and Experiment}, 2019\penalty0 (12):\penalty0 124009, 2019.

\bibitem[Ensinger et~al.(2023)Ensinger, Ziesche, Rakitsch, Tiemann, and Trimpe]{ensinger2023combining}
Katharina Ensinger, Sebastian Ziesche, Barbara Rakitsch, Michael Tiemann, and Sebastian Trimpe.
\newblock Combining slow and fast: Complementary filtering for dynamics learning.
\newblock \emph{arXiv preprint arXiv:2302.13754}, 2023.

\bibitem[Galetzka et~al.(2021)Galetzka, Loukrezis, and De~Gersem]{galetzka2021data}
Armin Galetzka, Dimitrios Loukrezis, and Herbert De~Gersem.
\newblock Data-driven solvers for strongly nonlinear material response.
\newblock \emph{International Journal for Numerical Methods in Engineering}, 122\penalty0 (6):\penalty0 1538--1562, 2021.

\bibitem[Ganchev et~al.(2011)Ganchev, Kral, Oberguggenberger, and Wolbank]{ganchev2011sensorless}
Martin Ganchev, Christian Kral, Helmut Oberguggenberger, and Thomas Wolbank.
\newblock Sensorless rotor temperature estimation of permanent magnet synchronous motor.
\newblock In \emph{IECON 2011-37th Annual Conference of the IEEE Industrial Electronics Society}, pages 2018--2023. IEEE, 2011.

\bibitem[Goodfellow et~al.(2016)Goodfellow, Bengio, and Courville]{goodfellow2016deep}
Ian Goodfellow, Yoshua Bengio, and Aaron Courville.
\newblock \emph{Deep learning}.
\newblock MIT press, 2016.

\bibitem[Grasman(1987)]{grasman1987asymptotic}
Johan Grasman.
\newblock Asymptotic methods for relaxation oscillations and applications.
\newblock \emph{Applied Mathematical Sciences}, 1987.

\bibitem[Greydanus et~al.(2019)Greydanus, Dzamba, and Yosinski]{greydanus2019hamiltonian}
Samuel Greydanus, Misko Dzamba, and Jason Yosinski.
\newblock Hamiltonian neural networks.
\newblock \emph{Advances in neural information processing systems}, 32, 2019.

\bibitem[Hamrock and Anderson(1983)]{hamrock1983rolling}
Bernard~J Hamrock and William~J Anderson.
\newblock Rolling-element bearings.
\newblock Technical report, 1983.

\bibitem[Hau{\ss}mann et~al.(2021)Hau{\ss}mann, Gerwinn, Look, Rakitsch, and Kandemir]{haussmann2021learning}
Manuel Hau{\ss}mann, Sebastian Gerwinn, Andreas Look, Barbara Rakitsch, and Melih Kandemir.
\newblock Learning partially known stochastic dynamics with empirical {PAC Bayes}.
\newblock In \emph{International Conference on Artificial Intelligence and Statistics}, pages 478--486. PMLR, 2021.

\bibitem[Hershey et~al.(2017)Hershey, Chaudhuri, Ellis, Gemmeke, Jansen, Moore, Plakal, Platt, Saurous, Seybold, et~al.]{hershey2017cnn}
Shawn Hershey, Sourish Chaudhuri, Daniel~PW Ellis, Jort~F Gemmeke, Aren Jansen, R~Channing Moore, Manoj Plakal, Devin Platt, Rif~A Saurous, Bryan Seybold, et~al.
\newblock {CNN} architectures for large-scale audio classification.
\newblock In \emph{2017 ieee international conference on acoustics, speech and signal processing (icassp)}, pages 131--135. IEEE, 2017.

\bibitem[Hilborn and Mangel(2013)]{hilborn2013ecological}
Ray Hilborn and Marc Mangel.
\newblock \emph{The ecological detective: confronting models with data (MPB-28)}.
\newblock Princeton University Press, 2013.

\bibitem[Hochreiter and Schmidhuber(1997)]{hochreiter1997long}
Sepp Hochreiter and J{\"u}rgen Schmidhuber.
\newblock Long short-term memory.
\newblock \emph{Neural computation}, 9\penalty0 (8):\penalty0 1735--1780, 1997.

\bibitem[Huang et~al.(2019)Huang, Chen, Yao, and He]{huang2019ecg}
Jingshan Huang, Binqiang Chen, Bin Yao, and Wangpeng He.
\newblock {ECG} arrhythmia classification using {STFT-based} spectrogram and convolutional neural network.
\newblock \emph{IEEE access}, 7:\penalty0 92871--92880, 2019.

\bibitem[Jazwinski(2007)]{jazwinski2007stochastic}
Andrew~H Jazwinski.
\newblock \emph{Stochastic processes and filtering theory}.
\newblock Courier Corporation, 2007.

\bibitem[Julier and Uhlmann(2004)]{julier2004unscented}
Simon~J Julier and Jeffrey~K Uhlmann.
\newblock Unscented filtering and nonlinear estimation.
\newblock \emph{Proceedings of the IEEE}, 92\penalty0 (3):\penalty0 401--422, 2004.

\bibitem[K\'alm\'an(1960)]{kalman1960new}
Rudolph~Emil K\'alm\'an.
\newblock A new approach to linear filtering and prediction problems.
\newblock 1960.

\bibitem[Karpatne et~al.(2017)Karpatne, Atluri, Faghmous, Steinbach, Banerjee, Ganguly, Shekhar, Samatova, and Kumar]{karpatne2017theory}
Anuj Karpatne, Gowtham Atluri, James~H Faghmous, Michael Steinbach, Arindam Banerjee, Auroop Ganguly, Shashi Shekhar, Nagiza Samatova, and Vipin Kumar.
\newblock Theory-guided data science: A new paradigm for scientific discovery from data.
\newblock \emph{IEEE Transactions on knowledge and data engineering}, 29\penalty0 (10):\penalty0 2318--2331, 2017.

\bibitem[Kurz et~al.(2022)Kurz, De~Gersem, Galetzka, Klaedtke, Liebsch, Loukrezis, Russenschuck, and Schmidt]{kurz2022hybrid}
Stefan Kurz, Herbert De~Gersem, Armin Galetzka, Andreas Klaedtke, Melvin Liebsch, Dimitrios Loukrezis, Stephan Russenschuck, and Manuel Schmidt.
\newblock Hybrid modeling: towards the next level of scientific computing in engineering.
\newblock \emph{Journal of Mathematics in Industry}, 12\penalty0 (1):\penalty0 1--12, 2022.

\bibitem[LeCun et~al.(1988)LeCun, Touresky, Hinton, and Sejnowski]{lecun1988theoretical}
Yann LeCun, D~Touresky, G~Hinton, and T~Sejnowski.
\newblock A theoretical framework for back-propagation.
\newblock In \emph{Proceedings of the 1988 connectionist models summer school}, volume~1, pages 21--28. San Mateo, CA, USA, 1988.

\bibitem[Liu and Wang(2019)]{liu2019multi}
Dehao Liu and Yan Wang.
\newblock Multi-fidelity physics-constrained neural network and its application in materials modeling.
\newblock \emph{Journal of Mechanical Design}, 141\penalty0 (12), 2019.

\bibitem[Liu et~al.(2009)Liu, Kuo, and Zhou]{liu2009virtual}
Lichuan Liu, Sen~M Kuo, and MengChu Zhou.
\newblock Virtual sensing techniques and their applications.
\newblock In \emph{2009 International Conference on Networking, Sensing and Control}, pages 31--36. IEEE, 2009.

\bibitem[Long et~al.(2018{\natexlab{a}})Long, She, and Mukhopadhyay]{long2018hybridnet}
Yun Long, Xueyuan She, and Saibal Mukhopadhyay.
\newblock Hybridnet: integrating model-based and data-driven learning to predict evolution of dynamical systems.
\newblock In \emph{Conference on Robot Learning}, pages 551--560. PMLR, 2018{\natexlab{a}}.

\bibitem[Long et~al.(2018{\natexlab{b}})Long, Lu, Ma, and Dong]{long2018pde}
Zichao Long, Yiping Lu, Xianzhong Ma, and Bin Dong.
\newblock {PDE-Net}: Learning {PDEs} from data.
\newblock In \emph{International conference on machine learning}, pages 3208--3216. PMLR, 2018{\natexlab{b}}.

\bibitem[Long et~al.(2019)Long, Lu, and Dong]{long2019pde}
Zichao Long, Yiping Lu, and Bin Dong.
\newblock {PDE-Net 2.0}: Learning {PDEs} from data with a numeric-symbolic hybrid deep network.
\newblock \emph{Journal of Computational Physics}, 399:\penalty0 108925, 2019.

\bibitem[Patel et~al.(2019)Patel, Rambach, Visentin, Rusev, Pfeiffer, and Yang]{patel2019deep}
Kanil Patel, Kilian Rambach, Tristan Visentin, Daniel Rusev, Michael Pfeiffer, and Bin Yang.
\newblock Deep learning-based object classification on automotive radar spectra.
\newblock In \emph{2019 IEEE Radar Conference (RadarConf)}, pages 1--6. IEEE, 2019.

\bibitem[Qiu et~al.(2021)Qiu, Pfrommer, Kloft, Mandt, and Rudolph]{qiu2021neural}
Chen Qiu, Timo Pfrommer, Marius Kloft, Stephan Mandt, and Maja Rudolph.
\newblock Neural transformation learning for deep anomaly detection beyond images.
\newblock In \emph{International Conference on Machine Learning}, pages 8703--8714. PMLR, 2021.

\bibitem[Rabiner and Schafer(2007)]{RabinerTheoryAA}
Lawrence~R. Rabiner and Ronald~W. Schafer.
\newblock Introduction to digital speech processing.
\newblock Now Publishers, Inc., 2007.

\bibitem[Raissi et~al.(2019)Raissi, Perdikaris, and Karniadakis]{raissi2019physics}
Maziar Raissi, Paris Perdikaris, and George~E Karniadakis.
\newblock Physics-informed neural networks: A deep learning framework for solving forward and inverse problems involving nonlinear partial differential equations.
\newblock \emph{Journal of Computational physics}, 378:\penalty0 686--707, 2019.

\bibitem[Reichstein et~al.(2019)Reichstein, Camps-Valls, Stevens, Jung, Denzler, and Carvalhais]{reichstein2019deep}
Markus Reichstein, Gustau Camps-Valls, Bjorn Stevens, Martin Jung, Joachim Denzler, and Nuno Carvalhais.
\newblock Deep learning and process understanding for data-driven earth system science.
\newblock \emph{Nature}, 566\penalty0 (7743):\penalty0 195--204, 2019.

\bibitem[Sadoughi and Hu(2019)]{sadoughi2019physics}
Mohammadkazem Sadoughi and Chao Hu.
\newblock Physics-based convolutional neural network for fault diagnosis of rolling element bearings.
\newblock \emph{IEEE Sensors Journal}, 19\penalty0 (11):\penalty0 4181--4192, 2019.

\bibitem[S{\"a}rkk{\"a} and Svensson(2023)]{sarkka2023bayesian}
Simo S{\"a}rkk{\"a} and Lennart Svensson.
\newblock \emph{Bayesian filtering and smoothing}, volume~17.
\newblock Cambridge university press, 2023.

\bibitem[Schapire(1990)]{schapire1990strength}
Robert~E Schapire.
\newblock The strength of weak learnability.
\newblock \emph{Machine learning}, 5:\penalty0 197--227, 1990.

\bibitem[Schirmer et~al.(2022)Schirmer, Eltayeb, Lessmann, and Rudolph]{schirmer2022modeling}
Mona Schirmer, Mazin Eltayeb, Stefan Lessmann, and Maja Rudolph.
\newblock Modeling irregular time series with continuous recurrent units.
\newblock In \emph{International Conference on Machine Learning}, pages 19388--19405. PMLR, 2022.

\bibitem[Stewart and Ermon(2017)]{stewart2017label}
Russell Stewart and Stefano Ermon.
\newblock Label-free supervision of neural networks with physics and domain knowledge.
\newblock In \emph{Thirty-First AAAI Conference on Artificial Intelligence}, 2017.

\bibitem[Stokes et~al.(2020)Stokes, Yang, Swanson, Jin, Cubillos-Ruiz, Donghia, MacNair, French, Carfrae, Bloom-Ackermann, et~al.]{stokes2020deep}
Jonathan~M Stokes, Kevin Yang, Kyle Swanson, Wengong Jin, Andres Cubillos-Ruiz, Nina~M Donghia, Craig~R MacNair, Shawn French, Lindsey~A Carfrae, Zohar Bloom-Ackermann, et~al.
\newblock A deep learning approach to antibiotic discovery.
\newblock \emph{Cell}, 180\penalty0 (4):\penalty0 688--702, 2020.

\bibitem[Thompson and Kramer(1994)]{thompson1994modeling}
Michael~L Thompson and Mark~A Kramer.
\newblock Modeling chemical processes using prior knowledge and neural networks.
\newblock \emph{AIChE Journal}, 40\penalty0 (8):\penalty0 1328--1340, 1994.

\bibitem[Toth et~al.(2019)Toth, Rezende, Jaegle, Racani{\`e}re, Botev, and Higgins]{toth2019hamiltonian}
Peter Toth, Danilo~Jimenez Rezende, Andrew Jaegle, S{\'e}bastien Racani{\`e}re, Aleksandar Botev, and Irina Higgins.
\newblock Hamiltonian generative networks.
\newblock \emph{arXiv preprint arXiv:1909.13789}, 2019.

\bibitem[Trimpe and D'Andrea(2010)]{trimpe2010accelerometer}
Sebastian Trimpe and Raffaello D'Andrea.
\newblock Accelerometer-based tilt estimation of a rigid body with only rotational degrees of freedom.
\newblock In \emph{2010 IEEE International Conference on Robotics and Automation}, pages 2630--2636. IEEE, 2010.

\bibitem[von Rueden et~al.(2019)von Rueden, Mayer, Beckh, Georgiev, Giesselbach, Heese, Kirsch, Pfrommer, Pick, Ramamurthy, et~al.]{von2019informed}
Laura von Rueden, Sebastian Mayer, Katharina Beckh, Bogdan Georgiev, Sven Giesselbach, Raoul Heese, Birgit Kirsch, Julius Pfrommer, Annika Pick, Rajkumar Ramamurthy, et~al.
\newblock Informed machine learning--a taxonomy and survey of integrating knowledge into learning systems.
\newblock \emph{arXiv preprint arXiv:1903.12394}, 2019.

\bibitem[Von~Stosch et~al.(2014)Von~Stosch, Oliveira, Peres, and de~Azevedo]{stosch2014hybrid}
Moritz Von~Stosch, Rui Oliveira, Joana Peres, and Sebasti{\~a}o~Feyo de~Azevedo.
\newblock Hybrid semi-parametric modeling in process systems engineering: Past, present and future.
\newblock \emph{Computers \& Chemical Engineering}, 60:\penalty0 86--101, 2014.

\bibitem[Wang et~al.(2017)Wang, Wu, and Xiao]{wang2017physic}
Jian-Xun Wang, Jin-Long Wu, and Heng Xiao.
\newblock Physics-informed machine learning approach for reconstructing {Reynolds} stress modeling discrepancies based on {DNS} data.
\newblock \emph{Physical Review Fluids}, 2\penalty0 (3):\penalty0 034603, 2017.

\bibitem[Willard et~al.(2022)Willard, Jia, Xu, Steinbach, and Kumar]{willard2022integrating}
Jared Willard, Xiaowei Jia, Shaoming Xu, Michael Steinbach, and Vipin Kumar.
\newblock Integrating scientific knowledge with machine learning for engineering and environmental systems.
\newblock \emph{ACM Computing Surveys}, 55\penalty0 (4):\penalty0 1--37, 2022.

\bibitem[Williams and Rasmussen(2006)]{williams2006gaussian}
Christopher~KI Williams and Carl~Edward Rasmussen.
\newblock \emph{Gaussian processes for machine learning}, volume~2.
\newblock MIT press Cambridge, MA, 2006.

\bibitem[Wilson et~al.(2016)Wilson, Hu, Salakhutdinov, and Xing]{wilson2016deep}
Andrew~Gordon Wilson, Zhiting Hu, Ruslan Salakhutdinov, and Eric~P Xing.
\newblock Deep kernel learning.
\newblock In \emph{Artificial intelligence and statistics}, pages 370--378. PMLR, 2016.

\bibitem[Wolpert(1992)]{WOLPERT1992241}
David~H. Wolpert.
\newblock Stacked generalization.
\newblock \emph{Neural Networks}, 5\penalty0 (2):\penalty0 241--259, 1992.
\newblock ISSN 0893-6080.
\newblock \doi{https://doi.org/10.1016/S0893-6080(05)80023-1}.
\newblock URL \url{https://www.sciencedirect.com/science/article/pii/S0893608005800231}.

\bibitem[Xu and Valocchi(2015)]{xu2015data}
Tianfang Xu and Albert~J Valocchi.
\newblock Data-driven methods to improve baseflow prediction of a regional groundwater model.
\newblock \emph{Computers \& Geosciences}, 85:\penalty0 124--136, 2015.

\bibitem[Yin et~al.(2021)Yin, Le~Guen, Dona, de~B{\'e}zenac, Ayed, Thome, and Gallinari]{yin2021augmenting}
Yuan Yin, Vincent Le~Guen, J{\'e}r{\'e}mie Dona, Emmanuel de~B{\'e}zenac, Ibrahim Ayed, Nicolas Thome, and Patrick Gallinari.
\newblock Augmenting physical models with deep networks for complex dynamics forecasting.
\newblock \emph{Journal of Statistical Mechanics: Theory and Experiment}, 2021\penalty0 (12):\penalty0 124012, 2021.

\end{thebibliography}
\end{document}